\pdfoutput=1
\documentclass[runningheads]{llncs}
\usepackage{booktabs}
\usepackage{amsmath,amsfonts,amssymb}
\usepackage{textcomp}
\usepackage{xcolor}
\usepackage{tcolorbox}
\usepackage{colortbl}

\usepackage[utf8]{inputenc} 
\usepackage[T1]{fontenc}    
\usepackage{hyperref} 
\usepackage{url}            
\usepackage{tabularx,booktabs}       
\usepackage{nicefrac}       
\usepackage{balance}
\usepackage{caption}

\usepackage{xspace}
\usepackage{color}
\usepackage{makecell}

\usepackage{algorithm}
\usepackage{algpseudocode}
\usepackage{algorithmicx}

\usepackage{xspace}
\usepackage{cleveref}
\usepackage{wrapfig}
\usepackage{enumitem}

\usepackage{framed}
\newenvironment{result}{\begin{framed}\centering\it}{\end{framed}}

\usepackage{multirow}
\def\BibTeX{{\rm B\kern-.05em{\sc i\kern-.025em b}\kern-.08em
    T\kern-.1667em\lower.7ex\hbox{E}\kern-.125emX}}
    
\usepackage{siunitx}
\usepackage{pifont}

\let\emptyset\varnothing

\algnewcommand{\LineComment}[1]{\State\(\triangleright\) #1}

\newcommand{\revise}[1]{\textcolor{black}{#1}}

\newcommand{\SMLT}{\textsc{AequeVox}\xspace}

\begin{document}
\title{AequeVox: Automated Fairness Testing of Speech Recognition Systems}
\titlerunning{AequeVox}
%
\author{First Author\inst{1}\orcidID{0000-1111-2222-3333} \and
Second Author\inst{2,3}\orcidID{1111-2222-3333-4444} \and
Third Author\inst{3}\orcidID{2222--3333-4444-5555}}
\author{Sai~Sathiesh~Rajan\inst{1}, 
Sakshi~Udeshi\inst{1},Sudipta~Chattopadhyay\inst{1}}
\authorrunning{S. Rajan et al.}
%
\institute{Singapore University of Technology and Design}
\maketitle              
\begin{abstract}
Automatic Speech Recognition (ASR) systems have become ubiquitous. They can be found in a variety of form factors and are increasingly important in our daily lives. As such, ensuring that these systems are equitable to different subgroups of the population is crucial.
In this paper, we introduce, \SMLT, an automated testing framework for evaluating the fairness of ASR systems. \revise{\SMLT simulates different environments to assess the effectiveness of ASR systems for different populations. In addition, we investigate whether the chosen simulations are comprehensible to humans. We further propose a fault localization technique capable of identifying words that are not robust to these varying environments.} Both components of \SMLT are able to operate in the absence of ground truth data.

We evaluated \SMLT on speech from four different datasets using three different commercial ASRs. Our experiments reveal that non-native English, female and Nigerian English speakers generate
{\em 109\%}, {\em 528.5\%} and {\em 156.9\%} more errors, on average than
native English, male and UK Midlands speakers, respectively. 
%
Our user study also reveals that 82.9\% of the simulations (employed through speech transformations) had a 
comprehensibility rating above seven (out of ten), with the lowest rating being 6.78. This further validates 
the fairness violations discovered by \SMLT. Finally, we show that the non-robust words, as predicted 
by the fault localization technique embodied in \SMLT, show {\em 223.8\%} more errors than the predicted 
robust words across all ASRs.

 
\end{abstract}

\section{Introduction}


Automated speech recognition (ASR) systems have made great 
strides in a variety of application areas e.g. smart home devices, robotics 
and handheld devices, among others.
\revise{
The wide variety of applications have made ASR systems serve 
increasingly diverse groups of people.
}
Consequently, it is crucial that such systems behave in 
a non-discriminatory fashion. 
%
This is particularly important because assistive technologies powered by 
ASR systems are often the primary mode of interaction 
for users with certain disabilities~\cite{assistiveSpeech}.
%
Consequently, it is 
critical that an ASR system employed in such systems is effective in 
diverse environments and across a wide variety of speakers 
(e.g. male, female, native English speakers, non-native English speakers)
\revise{since they are often deployed in safety-critical 
scenarios~\cite{goss2016incidence}.}

\begin{figure}[t]
\begin{center}
\includegraphics[scale=0.1]{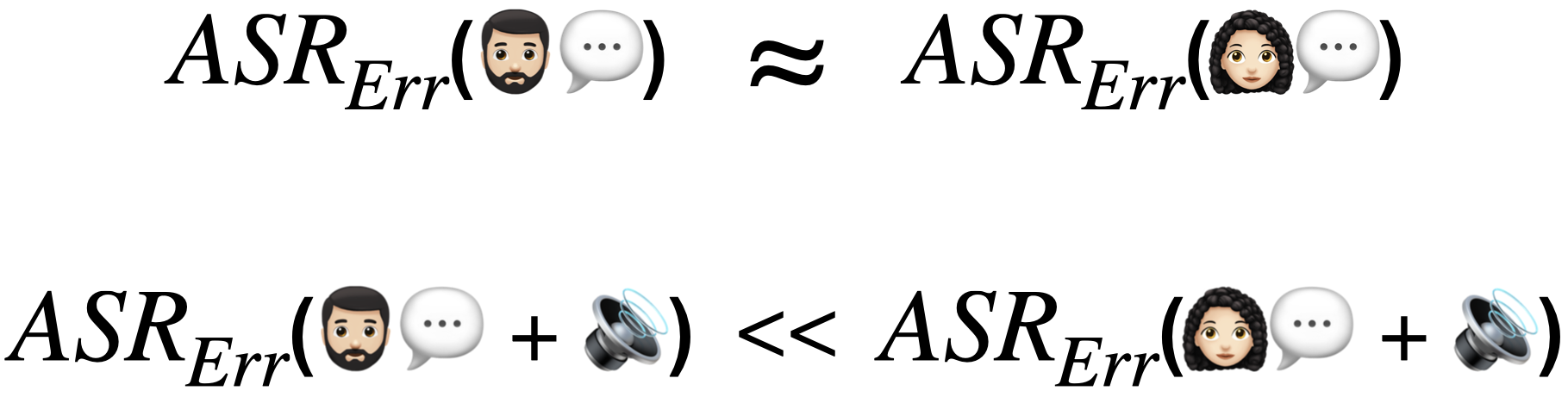}
\end{center}
\vspace{-0.2in}
\caption{Fairness Testing in \SMLT}
\label{fig:intro-diagram}
\vspace{-0.2in}
\end{figure}

In this paper, we are broadly concerned with the {\em fairness properties 
in ASR systems}. Specifically, {\em we investigate whether speech 
from one group is more robustly recognised as compared to another group}. 
For instance, consider the example shown in \Cref{fig:intro-diagram} for 
a system $\mathit{ASR}$. The metric $\mathit{ASR}_{\mathit{Err}}$ captures 
the error rate induced by $\mathit{ASR}$. Consider speech from two 
groups of speakers i.e. {\em male} and {\em female}. We assume that the 
$\mathit{ASR}$ has similar error rates for both the groups of speakers, 
as illustrated in the upper half of \Cref{fig:intro-diagram}. We now apply 
a small, constant perturbation on the speech provided by the two groups. 
Such a perturbation can be, for instance, addition of small noise, 
exemplifying the natural conditions that the ASR systems may need to work 
in (e.g. a noisy environment). If we observe that the $\mathit{ASR}_{\mathit{Err}}$ 
increases disproportionately for one of the speaker groups, as compared to 
the other, then we consider such a behaviour a violation of fairness (see 
the second half of \Cref{fig:intro-diagram}). 
\revise{Intuitively, \Cref{fig:intro-diagram} exemplifies the violations 
of {\em Equality of Outcomes}~\cite{phillips2004defending} in the context of ASR systems, 
where the male group is provided with a higher quality of service in a noisy 
environment as compared to the female group. Automatically discovering such 
scenarios of unfairness via simulating the ASR service in diverse 
environment is the main contribution of our \SMLT framework. }


%


\SMLT facilitates fairness testing without having any access to ground truth 
transcription data. 
Although, text-to-speech (TTS) can be used for generating speech, we argue that 
it is not suitable for accurately identifying the bias towards speech coming from 
a certain group.
%
Specifically, speakers may intentionally use enunciation, 
intonation, different degrees of loudness or other aspects of vocalization 
to articulate their message. Additionally, speakers unintentionally 
communicate their social characteristics such as their place of origin 
(through their accent), gender, age and education. This is unique 
to human speech and TTS systems cannot faithfully capture all the complexities 
inherent to human speech. 
%
%
Therefore, we believe that fairness testing of ASR 
systems should involve speech data from human speakers. 

We note that human speech (and the ASRs) may be subject to adverse environments (e.g. 
noise) and it is critical that the fairness evaluation considers such 
adverse environments. 
\revise{To facilitate the testing of ASR systems in adverse 
environments, we model the speech signal 
as a sinusoidal wave 
and subject it to eight different 
metamorphic transformations (e.g. noise, drop, low/high pass filter) that 
are highly relevant in real life.} Furthermore, in the absence of manually 
transcribed speech, we use a differential testing methodology to expose 
fairness violations.
In particular, \SMLT identifies the bias in ASR systems via a two step approach: 
Firstly, \SMLT registers the increase in error rates for speech from two groups 
when subjected to a metamorphic transformation. Subsequently, if the increase 
in the error rate of one group exceeds the other by a given threshold, \SMLT 
classifies this as a violation of fairness. 
To the best of our knowledge, we are unaware of any such differential testing 
methodology. \revise{As a by product of our \SMLT framework, we highlight words 
that contribute to errors by comparing the word counts from the original speech. 
This information can be further used to improve the ASR system.}

%
%
%




\revise{Existing works~\cite{themis,aequitas} isolate certain sensitive 
attributes (e.g. gender) and use such attributes to test for fairness. 
Isolating these attributes is
difficult in speech data, making it challenging to apply existing techniques to evaluate 
the fairness of ASR systems. \SMLT tackles this by
formalizing a unique fairness criteria targeted at ASR systems.}
%
Despite some existing efforts in testing ASR systems~\cite{crossASR,deepcruiser}, 
these  are not directly applicable for fairness testing. Additionally, some of these 
works require manually labelled speech transcription data~\cite{deepcruiser}. 
Finally, differential testing via TTS~\cite{crossASR} is not appropriate 
to determine the bias towards certain speakers, as they might use different 
vocalization that might be impossible (and perhaps irrational) to generate 
via a TTS. In contrast, \SMLT works on speech signals directly and defines 
transformations directly on these signals. \SMLT also does 
not require any access to manually labelled speech data for discovering 
fairness violations. 
%
%
In summary, we make the following contributions in the paper: 
\begin{enumerate}
%
%

\item \revise{We formalize a notion of fairness for ASR systems. This 
formalization draws parallels between the Equality of
Outcomes~\cite{phillips2004defending} and the quality of service provided 
by ASR systems in varying environments. }

\item \revise{We present \SMLT, which systematically combines 
metamorphic transformations and differential testing to highlight whether 
speech from a certain group (e.g. female) is subject to fairness violations 
by ASR systems. \SMLT neither requires access to ground truth transcription 
data nor does it require access to the ASR model structures}. 


\item We propose a fault localization method to identify the different words 
contributing to fairness errors.


\item We evaluate \SMLT with three different ASR systems namely Google Cloud, 
Microsoft Azure and IBM Watson. We use speech from the 
Speech Accent Archive~\cite{accents}, 
the Ryerson Audio-Visual Database of Emotional Speech and Song (RAVDESS)~\cite{ravdess}, 
Multi speaker Corpora of the English Accents in the British Isles 
(Midlands)~\cite{uk-midlands-dataset}, and a Nigerian English speech dataset~\cite{nigerian-english}. 
Our evaluation reveals that speech from non-native English speakers and female 
speakers \revise{exhibit higher fairness violations as compared 
to native English speakers and male speakers, respectively}. 

\item We validate the fault localization of \SMLT by showing that the identified 
{\em faulty} words generally introduce more errors to ASR systems even 
when used within speech generated via TTS systems. The inputs to the TTS system 
are randomly generated sentences that conform to a valid grammar. 

\item We evaluate (via the user study) the human comprehensibility score of the 
transformations employed by \SMLT on the speech signal. The lowest comprehensibility 
score was 6.78 and 82.9\% of the transformations had a comprehensibility score of more 
than seven. 
\end{enumerate}


\section{Background}
\label{sec:background}

In this section, we introduce the necessary background information. 

%

\smallskip \noindent
\textbf{Fairness in ASR Systems:}
\revise{A recent work, FairSpeech~\cite{racialASR}, 
uses conversational speech from black and white speakers 
to find that the word error rate for individuals who speak 
African American Vernacular English (AAVE) is nearly twice as large in all 
cases.}


\smallskip \noindent
\textbf{Testing ASR Systems:}
The \revise{major} testing focus, till date has been on image recognition systems and large language 
models. Few papers have probed ASR systems. One such work, Deep-Cruiser~\cite{deepcruiser} applies metamorphic transformations 
to audio samples to perform coverage-guided testing on ASR systems. 
Iwama et al.~\cite{automatedBasicRecognition} also perform automated testing on the basic recognition capabilities of ASR systems to detect functional defects.
CrossASR~\cite{crossASR} is another recent paper that applies differential testing 
to ASR systems. 

\smallskip \noindent
\textbf{The Gap in Testing ASR Systems:}
There is little work on automated methods to formalise and test fairness in ASR systems. 
In this work, we present \SMLT to test the fairness of ASR systems 
with respect to different population groups. It accomplishes this with the aid of differential testing 
of speech samples that have gone through metamorphic transformations of varying intensity. 
\revise{Our experimentation suggests that speech from different \revise{groups} of speakers receives significantly different quality of service across ASR systems.}
In the subsequent sections, we describe the design and evaluation of our \SMLT system.

\section{Methodology}
\label{sec:methodology}

In this section, we discuss \SMLT in detail. In particular, 
we motivate and formalize the notion of {\em fairness} in ASR systems. 
Then, we discuss our methodology to systematically find the violation of 
fairness in ASR systems. The notations used are described in 
\Cref{tab:Notation}.

\begin{table}[t]
\caption{Notations used}
\centering
\resizebox{\linewidth}{!}{
    \begin{tabular}{lp{14cm}}
        \toprule
        Notation & Description \\ \midrule
	    $GR_B$ & Base group  \\
	    $GR_k$ & $k \in (1, n)$. Various comparison group  \\
	    $MT$ & Metamorphic transformations \\
	    $ASR$ & Automatic Speech Recognition system under test \\
	    $\tau$ &  A user specified threshold beyond which the difference in 
	    word error rate for the base and comparison groups is considered 
	    a violation of individual fairness\\
        \bottomrule
    \end{tabular}}
    \label{tab:Notation}
\end{table}

\smallskip \noindent
\textbf{Motivation:} \revise{
Equality of outcomes~\cite{phillips2004defending} describes a state in which all people have approximately the same material wealth and income, or in which the general economic conditions of everyone's lives are alike. For a software system, equality of outcomes can be thought of as everyone getting the same quality of service from the software they are using. For a lot of software services, providing the same quality of service is baked into the system by design. For example, the results of a search engine only depend on the query. The quality of the result generally does not depend on any sensitive attributes such as race, age, gender and nationality. In the context of an ASR, the quality of service does depend on these sensitive attributes. This inferior quality of service may be especially detrimental in safety-critical settings such as emergency medicine~\cite{goss2016incidence} or air traffic management~\cite{kopald2013applying,helmke2016reducing}.}

\revise{
In our work, we show that the quality of service provided by ASR systems is vastly different depending on one’s gender/nationality/accent. Suppose there are two groups of people using an ASR system, males and females. They have approximately the same level of service when using this service at their homes. However, once they step into 
a different environment such as a noisy street, the quality of service drops notably for the female users, but 
does not drop noticeably for the male users. This is a violation of the principle of 
equality of outcomes (as seen for software systems) and more specifically, group 
fairness~\cite{dwork2012fairness}. Such a scenario is unfair (violation of group fairness) 
because some groups enjoy a higher quality of service than others.}

\revise{
In our work, we aim to automate the discovery of this unfairness. We do this by simulating the environment where 
the behaviour of ASR systems are likely to vary. The simulated environment is then enforced in speech from different groups. Finally, we measure how different groups are served in different environments.}

\smallskip \noindent
\textbf{Formalising Fairness in ASRs:}
In this section, we formalise the notion of fairness in the context of 
automated speech recognition systems (ASRs). The 
fairness definition in ASRs is as follows:
\begin{equation}
\begin{gathered}
	\left| \mathit{ASR}_{\mathit{Err}}(\mathit{GR}_{i}) - \mathit{ASR}_{\mathit{Err}}(\mathit{GR}_{j}) \right| \leq \tau 
	\label{eqn:simple-fairness}
\end{gathered}
\end{equation} 
Here, $\mathit{GR}_{i}$ and $\mathit{GR}_{j}$ capture speech from distinct groups 
of people. If the error rates induced by $\mathit{ASR}$ for group $\mathit{GR}_{i}$ 
($\mathit{ASR}_{\mathit{Err}}(\mathit{GR}_{i})$) and for group $\mathit{GR}_{j}$ 
($\mathit{ASR}_{\mathit{Err}}(\mathit{GR}_{j})$) differ beyond a certain threshold, 
we consider this scenario to be unfair. Such a notion of unfairness was studied in 
a recent work~\cite{racialASR}.

In this work, we want to explore \revise{whether different groups are fairly treated under varying conditions.}
\revise{Intuitively, we subject speech from different groups to a variety of 
simulated environments. We then measure the word error rates of the speech in such 
simulated environments and check if certain groups fare better than others. }
%
Formally, we capture the notion of fairness targeted by \SMLT as follows: 
\begin{equation}
\begin{gathered}
	D_{i} \gets  \mathit{ASR}_{\mathit{Err}}(\mathit{GR}_{i}) - 
	\mathit{ASR}_{\mathit{Err}}(\mathit{GR}_{i} + \delta)  \\
	D_{j} \gets  \mathit{ASR}_{\mathit{Err}}(\mathit{GR}_{j}) - 
	\mathit{ASR}_{\mathit{Err}}(\mathit{GR}_{j} + \delta) \\
	\left| D_{i} - D_{j}\right| \leq \tau 	
\label{eqn:robustness-fairness}
\end{gathered}
\end{equation} 
Here we perturb the speech of the two groups ($GR_{i}$ and $GR_{j}$) by 
adding some $\delta$ to the speech. We compare the degradation in the 
speech ($D_{i}$ and $D_{j}$). 
\revise{If the degradation faced by one group is far greater than the 
one faced by the other, we have a fairness violation. This is because  
speech from both groups ought to face similar degradation when subject 
to similar environments (simulated by $\delta$ perturbation) when 
equality of outcomes~\cite{phillips2004defending} holds. More 
specifically, this is a group fairness violation
because the quality of service (outcome) depends on 
the group~\cite{dwork2012fairness,fairnessDefinitions}.}


%
%
%
%

\begin{algorithm}[t]

    \caption{\SMLT Fairness Testing}
    {\scriptsize
    \begin{algorithmic}[1] 
        \Procedure{Fairness\_Testing}
        {$\mathit{GR}_B,\mathit{MT},\mathit{GR}_1, \cdots, \mathit{GR}_n,\tau, \mathit{ASR}_1, \mathit{ASR}_2$}
        	\State $Error\_Set \gets \emptyset$ 
        	
        	\For{$T \in \mathit{MT}$}
        	\State $\mathit{GR}_B^T \gets T(\mathit{GR}_B)$
        	\LineComment $\mathcal{L}$ computes the average 
        	word level levenshtein distance
        	\LineComment between the outputs of $\mathit{ASR}_1$ and $\mathit{ASR}_2$   
        	\State $d_B \gets \mathcal{L} (\mathit{ASR}_1(\mathit{GR}_B), \mathit{ASR}_2(\mathit{GR}_B))$
        	\State $d_B^T \gets \mathcal{L} (\mathit{ASR}_1(\mathit{GR}_B^T), \mathit{ASR}_2(\mathit{GR}_B^T))$
        	\State $D_B \gets d_B^T - d_B$

        		\For{$k \in (1, n)$}
        			\State $\mathit{GR}_k^T \gets T(\mathit{GR}_k)$
        			\State $d_k \gets \mathcal{L} ( \mathit{ASR}_1(\mathit{GR}_k), \mathit{ASR}_2(\mathit{GR}_k))$
        			\State $d_k^T \gets\mathcal{L} ( \mathit{ASR}_1(\mathit{GR}_k^T), \mathit{ASR}_2(\mathit{GR}_k^T))$
        			\State $D_k \gets d_k^T - d_k$
        			\If{$D_B - D_k > \tau$}
        				\State $Error\_Set \gets Error\_Set \cup 
        				(\mathit{GR}_B, \mathit{GR}_k, T)$
        			\EndIf
        		\EndFor
        	\EndFor
        	\State \Return $Error\_Set$
            
        \EndProcedure
    \end{algorithmic}
    }
    \label{alg:test-gen}
 \end{algorithm}

\smallskip \noindent
\textbf{Example:}
To motivate our system, let us sketch out an example. Consider texts of approximately 
the same length spoken by two sets of speakers whose native languages are 
$L_1$ and $L_2$ respectively. 
Let us assume that both sets of speakers read out a text in English. \SMLT uses 
two ASR systems and obtains the transcript of this speech. \SMLT then employs 
differential testing to find the word-level levenshtein distance~\cite{levenshtein-dist} 
between these two sets of transcripts. 
Let us also assume that the \revise{average} 
word-level levenshtein distance is two and four for 
$L_1$ and $L_2$ native speakers, respectively. 

\revise{
\SMLT then simulates a noisy environment by adding noise to the speech and 
obtains the transcript of this transformed speech.}
%
%
Let us assume now that the \revise{average} levenshtein distance
for this transformed speech is 4 and 25 for $L_1$ and $L_2$ native speakers, 
respectively. It is clear that the degradation for the speech of 
native $L_2$ speakers is much more severe. \revise{In this case, the quality of 
service that $L_2$ native speakers receive in noisy environments
is worse than $L_1$ native speakers. This is a violation of 
fairness which \SMLT aims to detect.} 

The working principle behind \SMLT holds even if the spoken text is different. 
This is because \SMLT just measures the relative degradation in ASR performance 
for a set of speakers. For large datasets, we are able to measure the average 
degradation in ASR performance with respect to different groups of speakers 
(e.g. {\em male}, {\em female}, {\em native, non-native English} speakers). 

\begin{figure*}[t]
\begin{center}
\resizebox{\linewidth}{!}{
\begin{tabular}{cccc}
\includegraphics[scale=0.25]{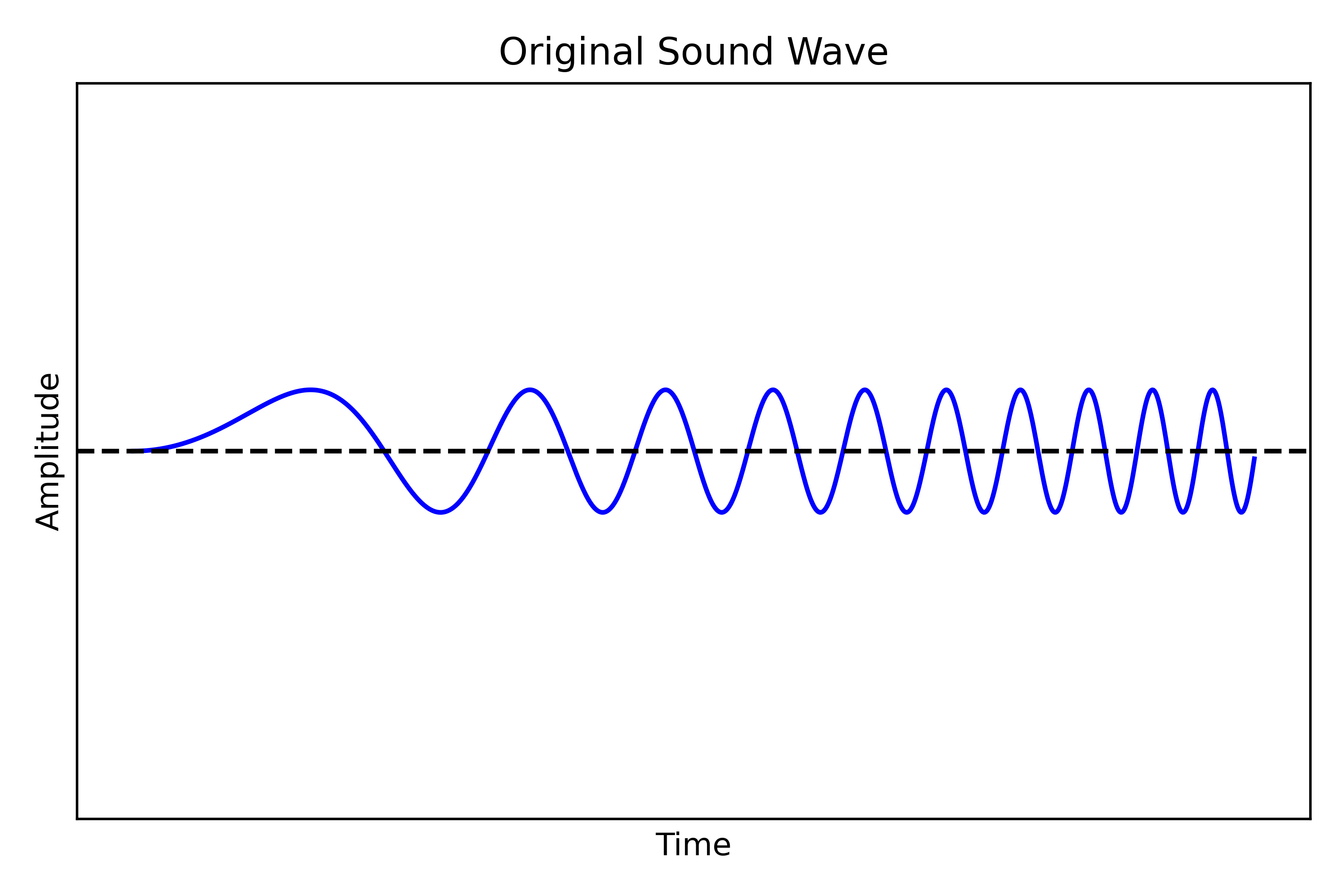} & 
\includegraphics[scale=0.25]{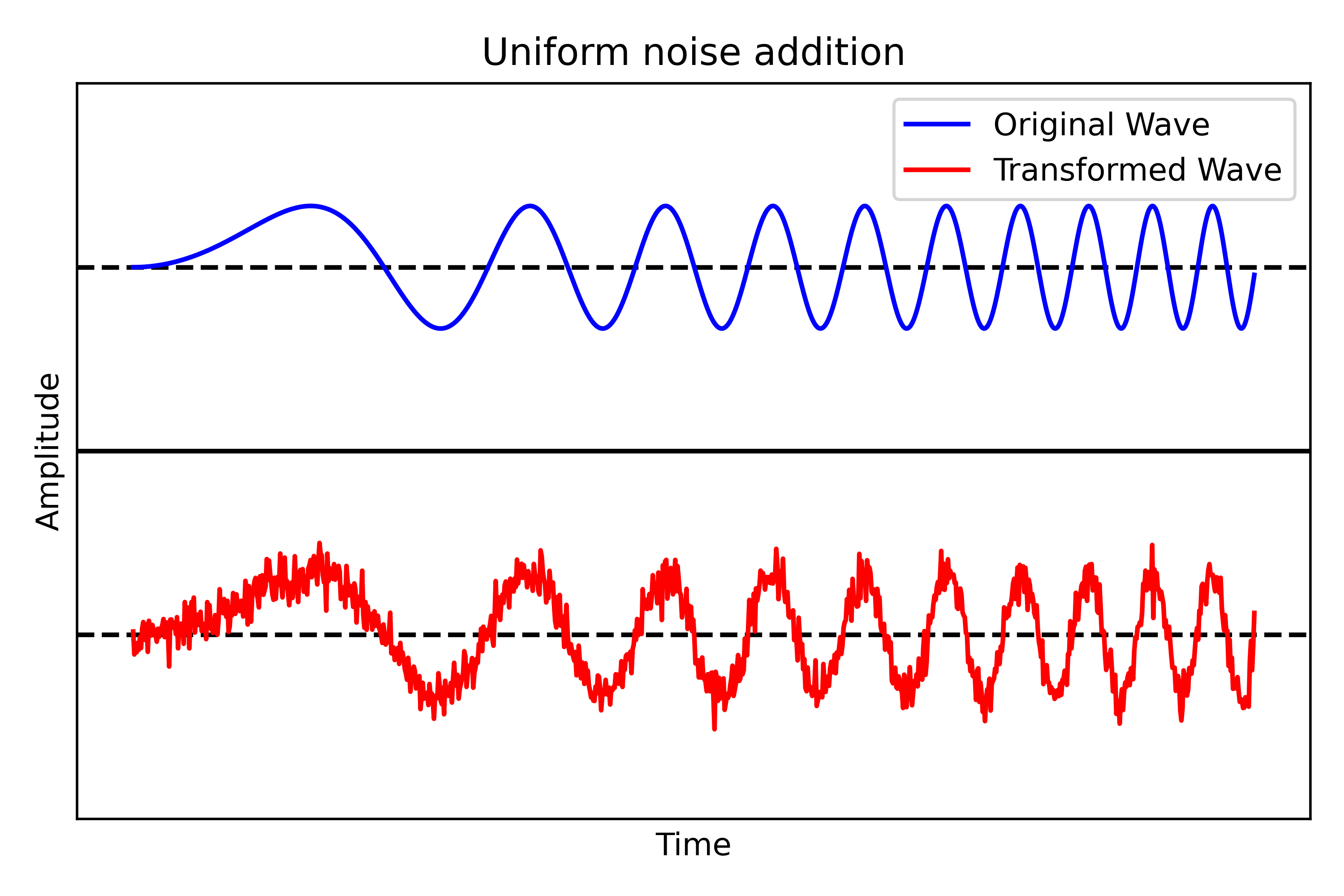} & 
\includegraphics[scale=0.25]{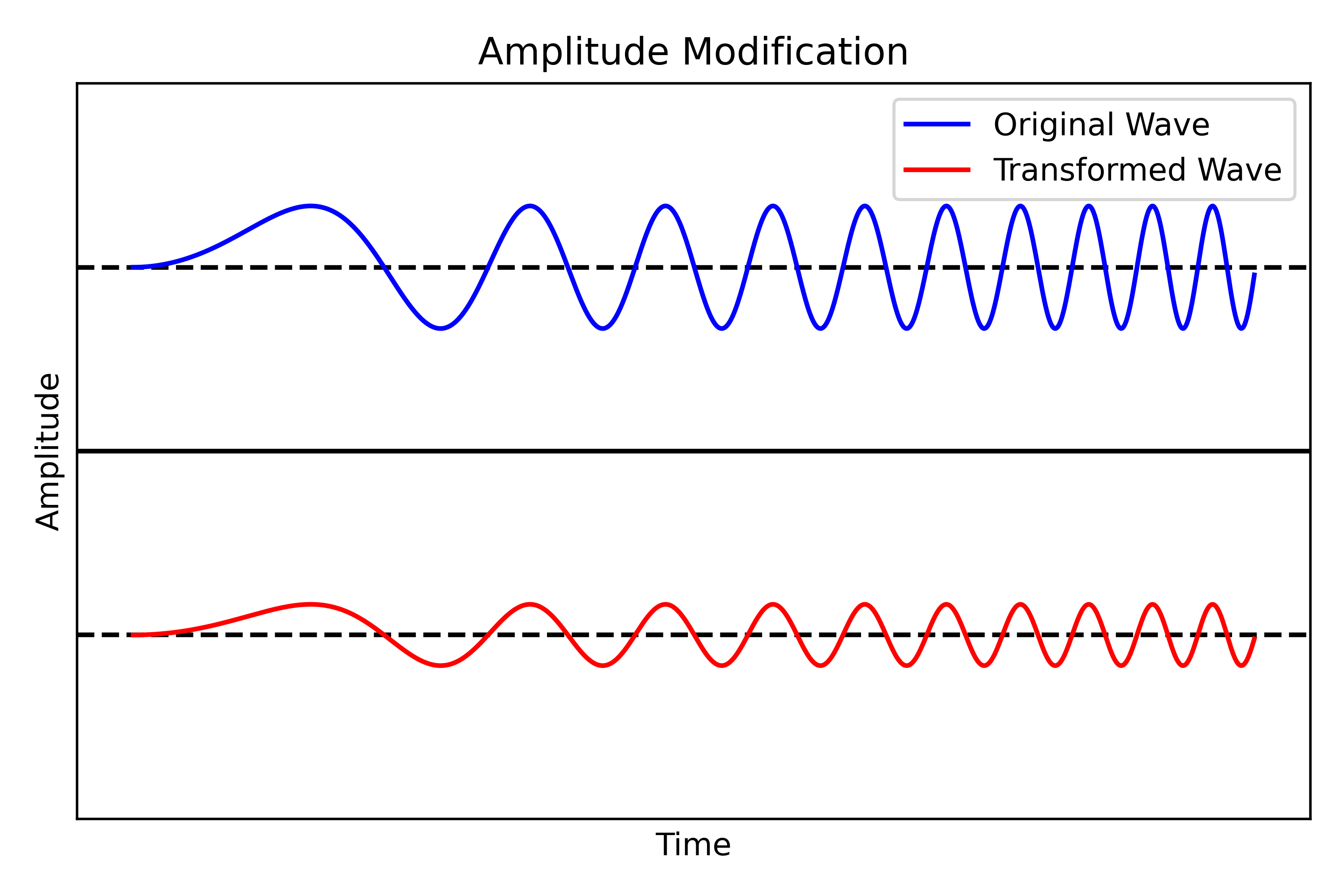} & 
\includegraphics[scale=0.25]{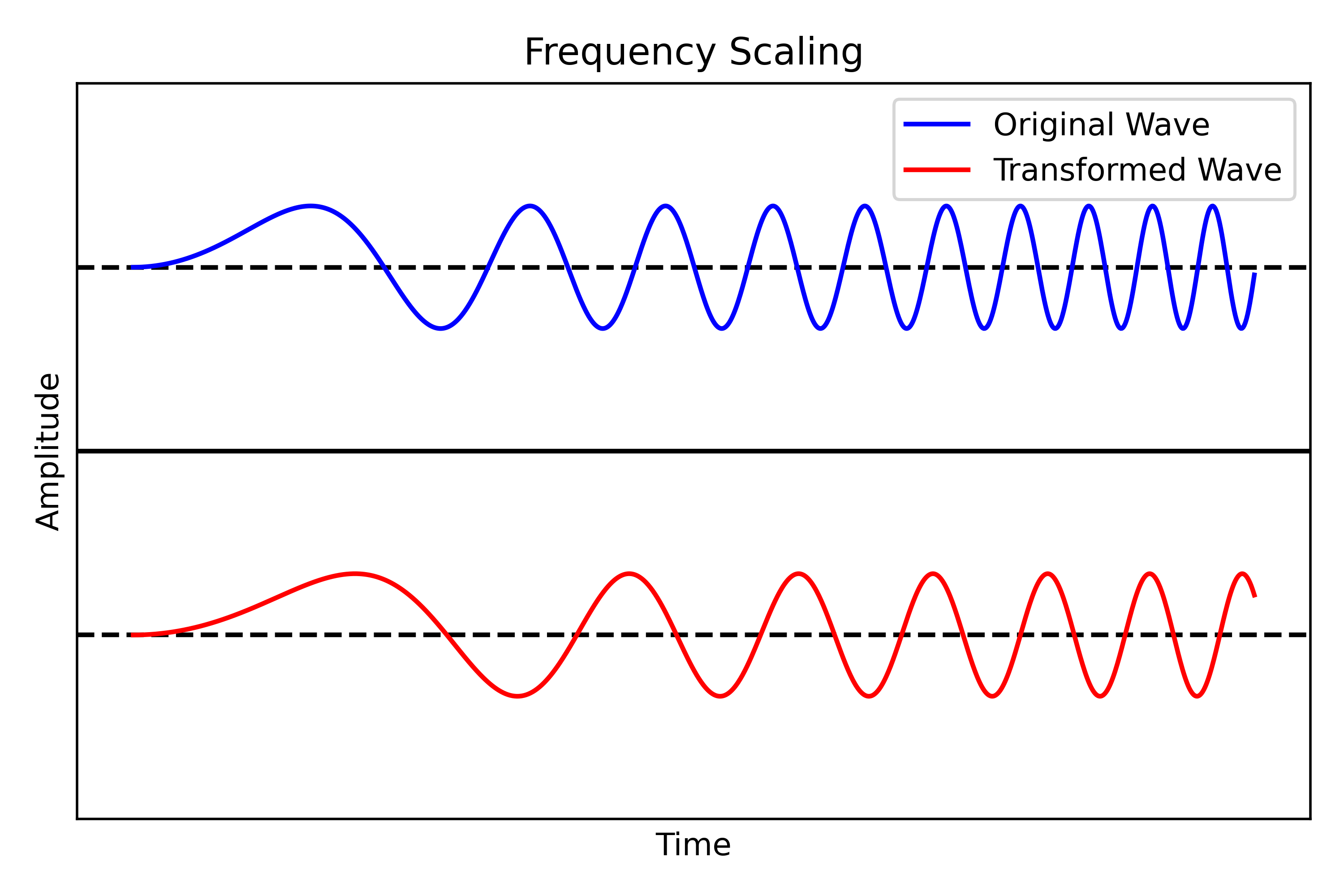}
 \\
{\bf (a)} & {\bf (b)} & {\bf (c)} & {\bf (d)}\\
\end{tabular}}
\\
\resizebox{\linewidth}{!}{
\begin{tabular}{cccc}
\includegraphics[scale=0.25]{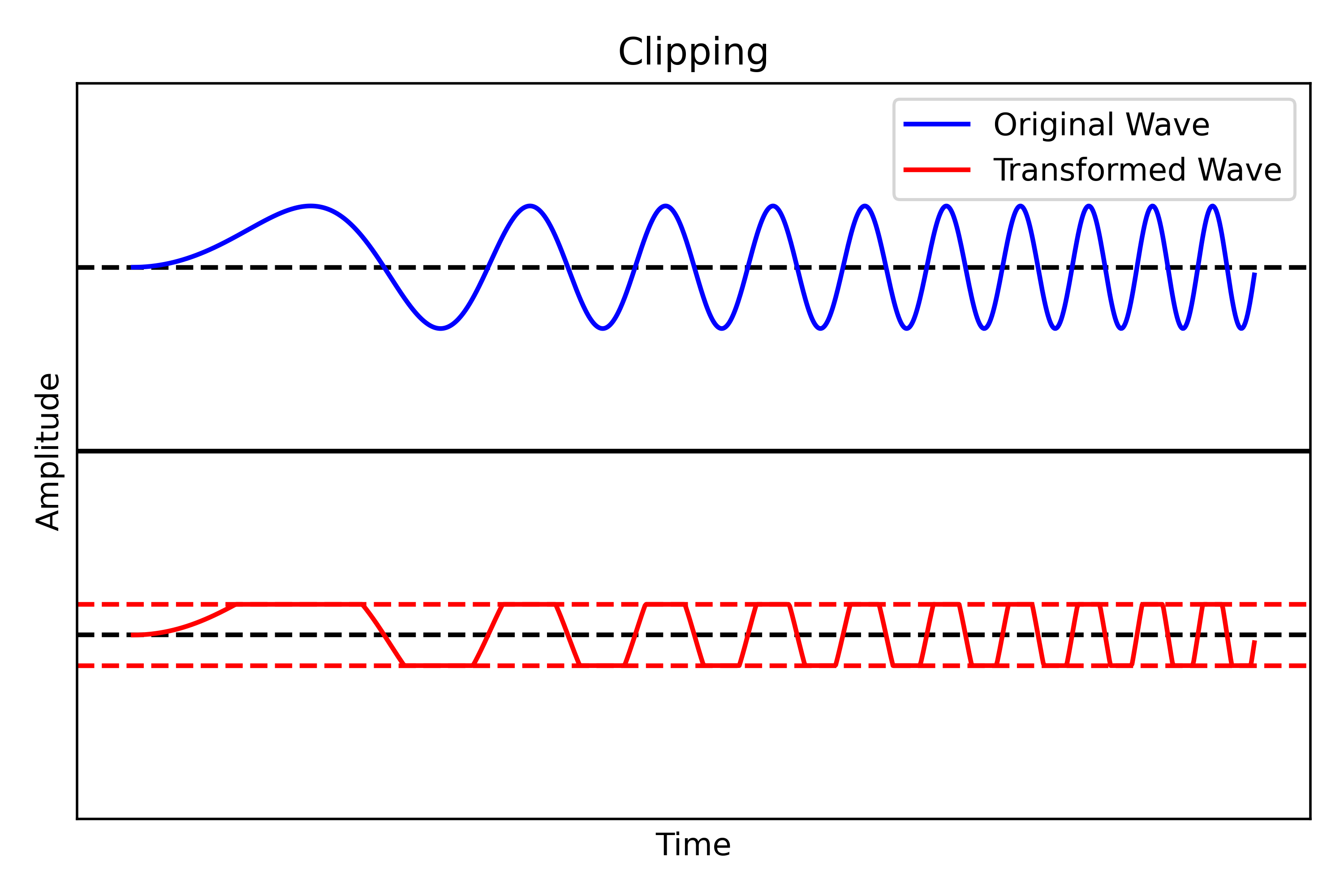} & 
\includegraphics[scale=0.25]{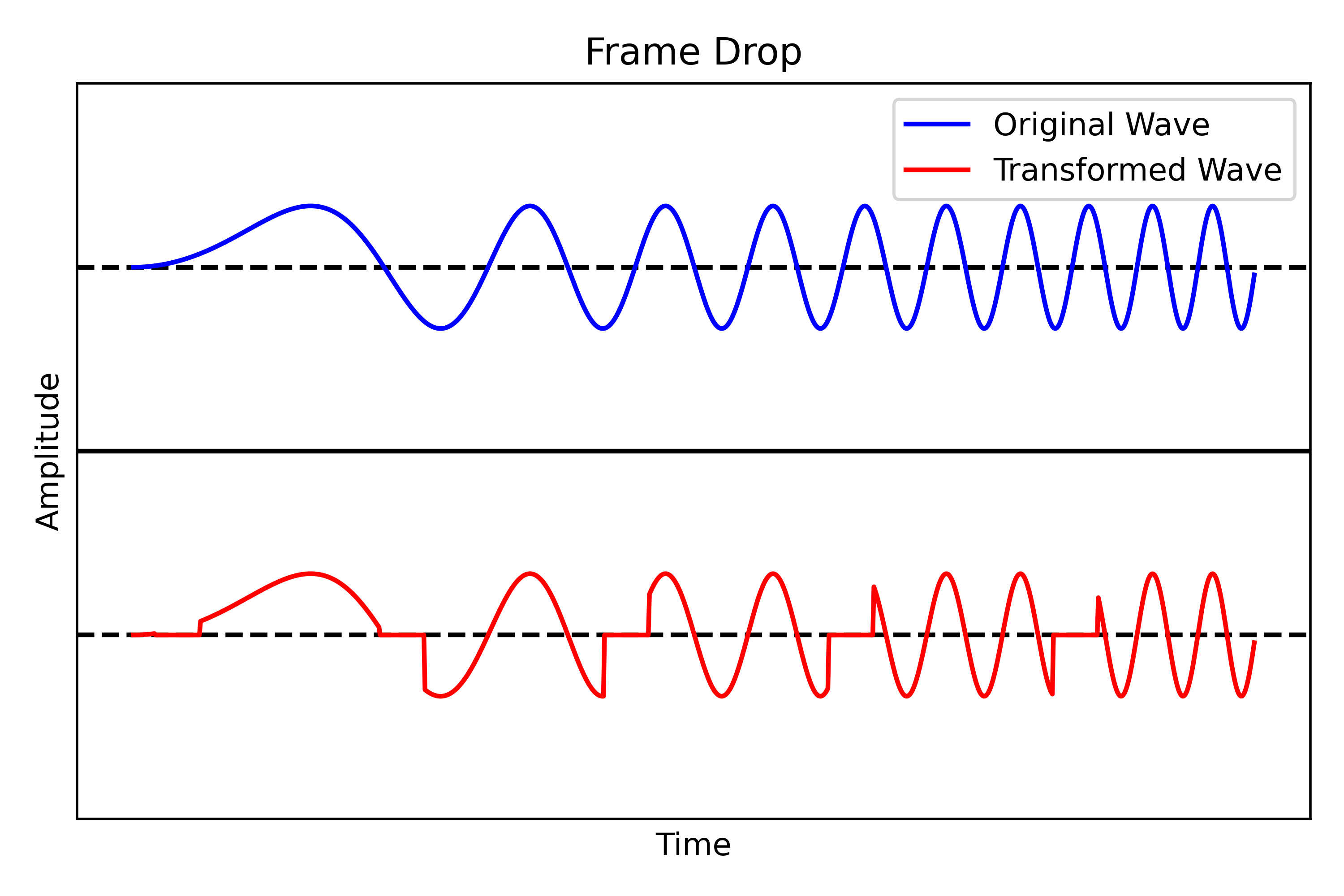} & 
\includegraphics[scale=0.25]{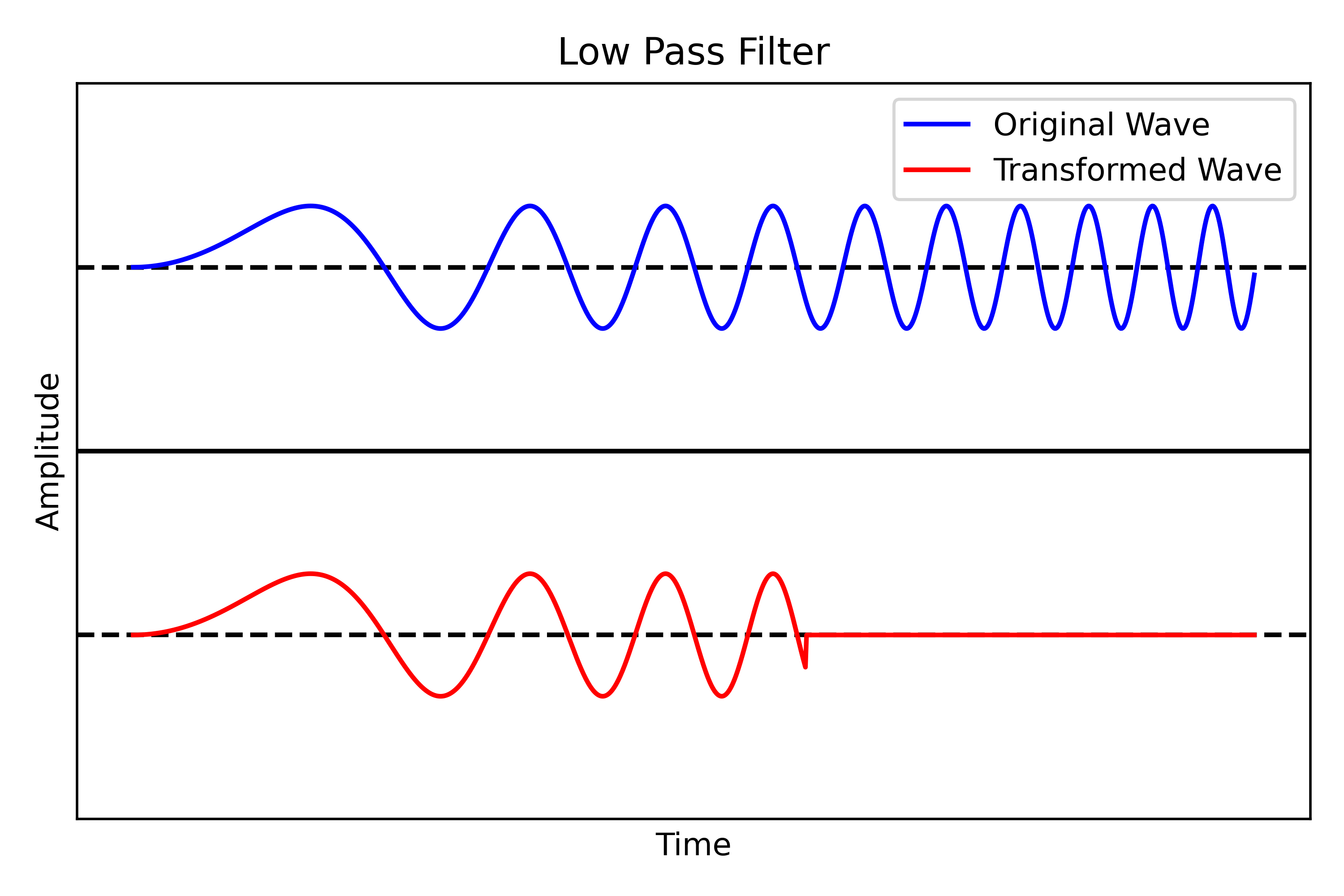} & 
\includegraphics[scale=0.25]{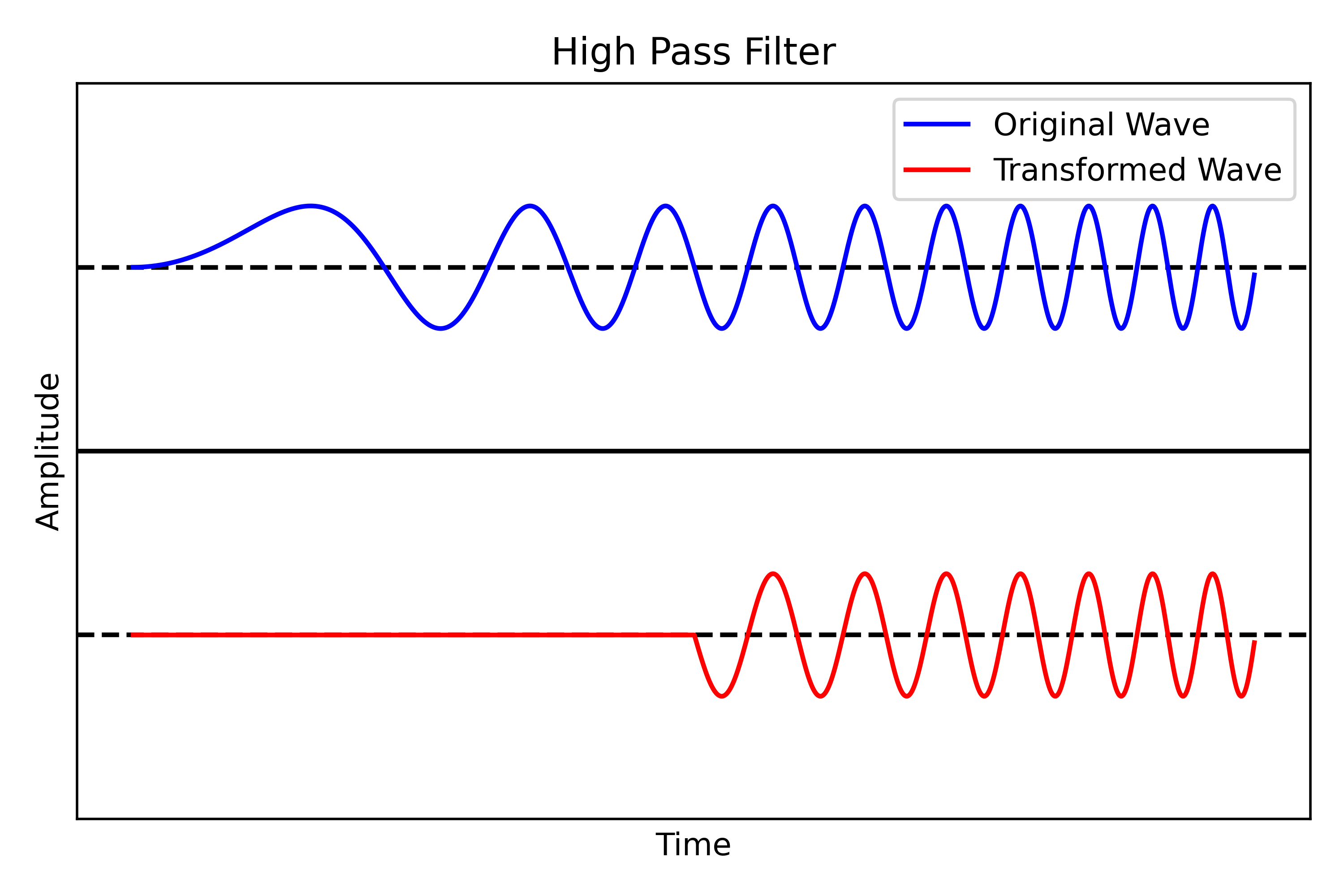} \\
{\bf (e)} & {\bf (f)} & {\bf (g)} & {\bf (h)} \\
\end{tabular}}
\end{center}
\vspace{-0.2in}
\caption{Sound wave transformations}
\label{fig:soundwave-transform}
\end{figure*}

\smallskip \noindent
\textbf{Metamorphic Transformations of Sound:} 
\label{sec:meta-trans-methodology}
\revise{
The ability to operate in a wide range of environments is crucial in ASR systems 
as they are deployed in safety-critical settings such as medical emergency 
services~\cite{goss2016incidence} and air traffic 
managment~\cite{helmke2016reducing}, \cite{kopald2013applying}, which are 
known to have interference and noise. Metamorphic speech transformations serve to
simulate such scenarios.
The key insight for our metamorphic transformations comes from how 
waves are represented and what can happen to these 
waves when they're transmitted in different mediums.} We realise 
this insight in the fairness testing system for ASR systems. To the 
best of our knowledge \SMLT is the first work that combines this insight 
from \revise{acoustics}, software testing and software fairness 
to evaluate the fairness of ASR systems. 
\revise{
\SMLT uses the addition of noise (\Cref{fig:soundwave-transform} (b)),
amplitude modification (\Cref{fig:soundwave-transform} (c)), 
frequency modification (\Cref{fig:soundwave-transform} (d)),
amplitude clipping (\Cref{fig:soundwave-transform} (e)),
frame drops (\Cref{fig:soundwave-transform} (f)), 
low-pass filters (\Cref{fig:soundwave-transform} (g)), and
high-pass filters (\Cref{fig:soundwave-transform} (h)) as metamorphic speech
transformations. We choose these transformations because they are the 
most common distortions for sound in various environments~\cite{kaggleAudio}. 
The details of the transformations are in
\Cref{sec:sound-transformations}.}

\begin{figure}[t]
\begin{center}
\includegraphics[scale=0.18]{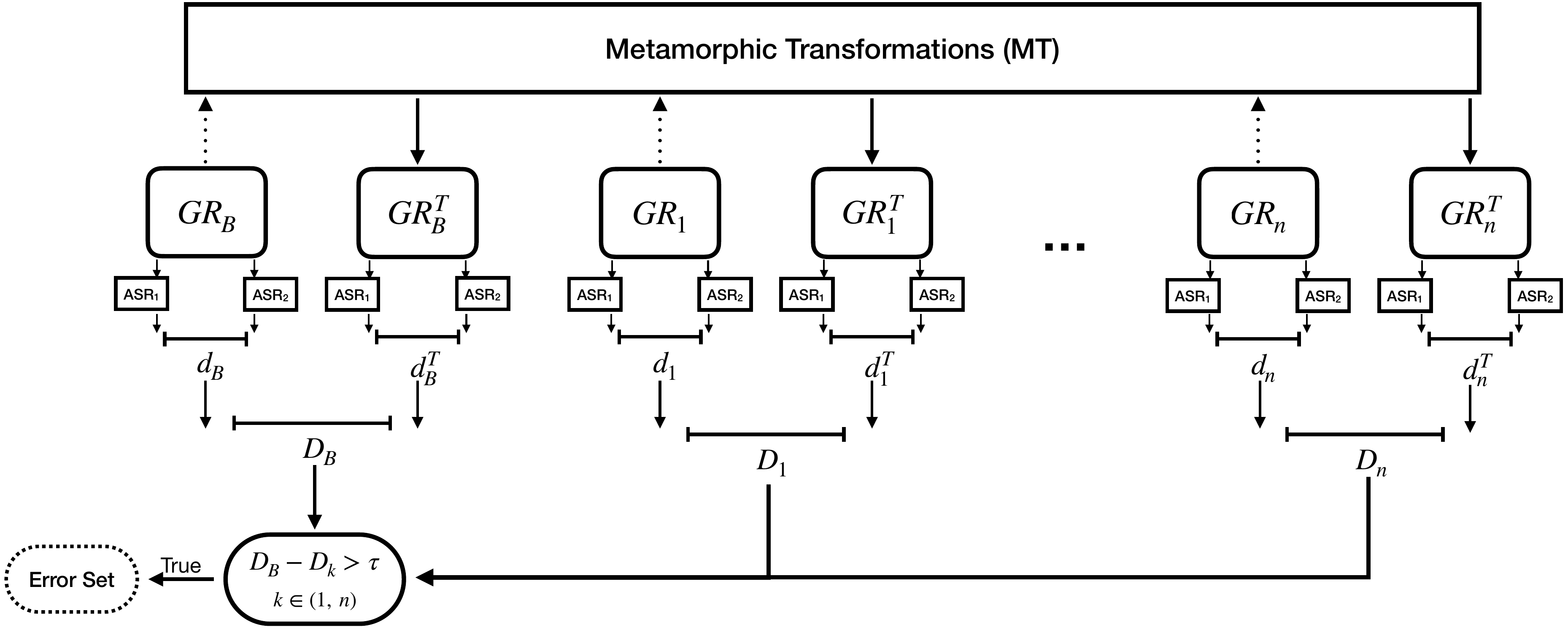}
\end{center}
\vspace{-0.2in}
\caption{\SMLT System Overview}
\label{fig:overview}
\vspace{-0.2in}
\end{figure}

\smallskip \noindent
\textbf{System Overview:}
Algorithm~\ref{alg:test-gen} provides an outline of our overall test generation process. 
We realise the notion of fairness described in 
\Cref{eqn:robustness-fairness} using differential testing. The error
rates ($\mathit{ASR}_{Err}$) for a particular speech \revise{clip}
are found by finding the difference 
between the outputs of two ASR systems, ASR$_1$ and ASR$_2$.
It is important to note that we make a design choice to use 
{\em differential testing} to find the error rate ($\mathit{ASR}_{Err}$). 
This helps us eliminate the need for ground truth transcription data which 
is both labor intensive and expensive to obtain.
Furthermore, \SMLT realises the $\delta$ seen in 
\Cref{eqn:robustness-fairness} by using metamorphic transformations 
for speech ({\em see \Cref{fig:overview}}). 
\revise{These speech metamorphic transformations represent the various simulated 
environments for which \SMLT wants to measure the quality of service for different
groups.}
Additionally, the user can customise this $\delta$ per their requirements.
In our implementation we use eight distinct metamorphic transformations
 as $\delta$ 
({\em see \Cref{fig:soundwave-transform}}).
Specifically, we investigate 
how fairly do two ASR systems (ASR$_1$ and ASR$_2$) treat groups
($GR_k^T \mid k \in \{1, 2, \cdots n\}$) with respect to a base group 
($GR_B$). 
\SMLT achieves this by taking a dataset of speech which contains 
data from two or more different groups (e.g. male and female speakers, Native 
English and Non-native English speakers) and modifies these speech snippets 
through a set of transformations ($MT$). These are then divided 
into base group transformed speech ($GR_B^T$) and the 
transformed speech for other groups 
($GR_k^T \mid k \in \{1, 2, \cdots n\}$). 
As seen in \Cref{alg:test-gen}, the average word-level levenshtein 
distance (word-level levenshtein distance divided by the number of words in the longer transcript) between the outputs of the two 
ASR systems is captured by $d_B$ and $d_B^T$ for the original and transformed
speech respectively. 
Similarly, for the comparison groups $GR_k^T (k \in 
\{1, 2, \cdots n\})$ the word-level levenshtein distance is captured by $d_k$ 
and $d_k^T$. The higher the levenshtein distance the larger the error in 
terms of differential testing. In other words, larger error in differential testing 
would mean that the ASR systems disagree on a higher number of words. 

To capture the \revise{degradation in the quality of service} 
for the speech \revise{subjected to simulated environments ($MT$)},
we compute the difference between 
the word-level levenshtein distance for the original and transformed 
speech. Specifically, we compute $D_B$ as $d_B^T - d_B$ and 
$D_k$ as $d_k^T - d_k (k \in 
\{1, 2, \cdots n\})$ for the base and comparison groups, respectively. 
The higher this metric ($D_B$ and $D_k$), the more severe
the degradation in ASR \revise{quality of service} 
because of the transformation~$T$. 

We compare these metrics and if $D_B$ exceeds $D_k$ by some 
threshold $\tau$, we classify this as an error 
\revise{for the {\em base group} ($GR_B$)} and more specifically a 
violation of fairness ({\em see \Cref{fig:overview}}). In our experiments
we set each of the groups in our dataset as the base group ($GR_B$) and
run the \SMLT technique to find errors with respect to that base group. 
The lower the errors (as computed via the violation of the assertion 
$D_B - D_k \leq \tau$), the fairer the ASR systems are with respect 
to groups $GR_B$. 
\revise{As an example, let us say Russian 
speakers are the base group ($GR_B$), English speakers are the 
comparison group ($GR_k$) and the value of $\tau$ is 0.1. If $D_B$ 
is strictly greater than $D_k$ by 0.1, then fairness violation is counted for the Russian 
speakers}. 
\revise{Otherwise, no fairness errors are recorded.}


%
%
%
%
%
%
%
%
%
%


\smallskip \noindent
\textbf{Fault Localisation:}
%
%
\revise{\SMLT introduces a word-level fault
localisation technique, which does not require any access to ground truth 
data}. We first illustrate a use case of this fault localisation technique. 

\smallskip \noindent
\textbf{Example:} Let us consider a corpus of English sentences by a group of 
speakers (say $GR$) who speak language $L_1$ natively. \SMLT builds a dictionary for all 
the words in the transcript obtained from $ASR_1$. An excerpt from such a dictionary 
appears as follows: \{\textit{brother}~:~16, \textit{nice}~:~25, 
\textit{is}~:~33,~$\cdots$\}. This means the words {\em brother, nice} and 
{\em is} were seen 16, 25 and 33 times in the transcript respectively. 
Now, assume \SMLT \revise{simulates a noisy environment}
\revise{
by adding noise with various signal to noise (SNR) ratios as follows: 
\{\textit{10, 8, 6, 4, 2}\}.} This is the parameter for the transformation ($param^T$).


Once \SMLT 
obtains the transcript of these transformed inputs, it creates 
dictionaries similar to the ones seen in the preceding paragraph. 
Let the relevant subset of the dictionary for SNR {\em two} (2) be 
\{\textit{brother}~:~1, \textit{nice}~:~23, \textit{is}~:~32,~$\cdots$\}. 
We use this to determine that the utterance
of the word {\em brother} is not robust for noise addition for the group
$GR$. This is because, the word brother appears significantly less in the 
transcript for the modified speech, as compared to the transcript for the 
original speech. 

\begin{algorithm}[t]
	\caption{\SMLT Fault Localizer}
    {\scriptsize
    \begin{algorithmic}[1]
        \Procedure{Fault\_Localizer}
        {$\mathit{WC}$, $\mathit{WC}^{T_{\theta}}$, $\omega$, $param^T$}
        	\State Drop\_Count $\gets \emptyset$ 
        	\State Non\_Robust\_Words $\gets \emptyset$ 
        	
        	\For{$word$ $\in \mathit{WC}.$keys()}
        		\State $init\_count \gets \mathit{WC}[word]$
        		\LineComment Returns the minimum count of $word$ across all 
        		the parameter 
        		\LineComment of transformation $T$ 
        		\State $min\_count \gets$ get\_min($\mathit{WC}^{T_{\theta}}[word]$, 
        		$~param^T$)
        		\State $\mathit{count\_diff} \gets$ max
        		(($init\_count - min\_count)$, 0) 
        		\If{$\mathit{count\_diff} > \omega$}
        			\State Non\_Robust\_Words $\gets$ Non\_Robust\_Words 
        			$~\cup~\{word\}$
        		\EndIf{}
        		
        		\State Drop\_Count $\gets$ Drop\_Count $~\cup~\{count\_diff\}$
        	\EndFor
        	
        	\State \Return {Non\_Robust\_Words, Drop\_Count}
 	\EndProcedure
    \end{algorithmic}
    }
    \label{alg:fault-loc}
 \end{algorithm}

\begin{figure}[t]
\begin{center}
\includegraphics[scale=0.52]{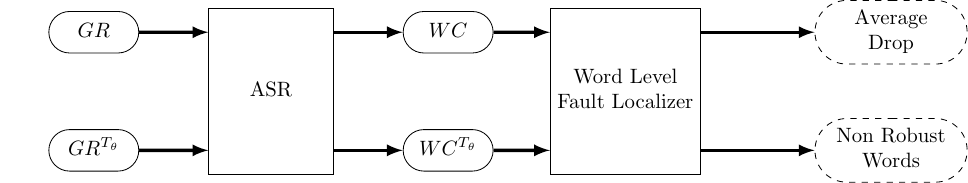}
\end{center}
\vspace{-0.2in}
\caption{\SMLT Fault Localization Overview}
\label{fig:fault-loc}
\vspace{-0.2in}
\end{figure}

\smallskip \noindent
\textbf{\SMLT fault localisation overview:}
Algorithm~\ref{alg:fault-loc} provides an overview of the fault localization 
technique implemented in \SMLT. 
The goal of the \SMLT fault localisation is to find words for
a group ($GR$) that are not robust to \revise{
the simulated environments. Specifically, \SMLT finds words which are not 
recognised by the ASR when subjected to the appropriate speech 
transformations.
}


The transformation 
is represented by $T_{\theta}$. Here, $T \in MT$ is the transformation and 
${\theta} \in param^T$ is the parameter of the transformation, which 
controls \revise{the severity of the transformation.}

As seen in \Cref{alg:fault-loc},
\SMLT builds a word count dictionary for each word in $WC$ and $WC^{T_{\theta}}$ for 
the original speech and for each ${\theta} \in param^T$ respectively. 
For each word, \SMLT finds the difference in the number of appearances for a 
word in $WC$ and in $WC^{T_{\theta}}$ for ${\theta} \in param^T$. To compute the difference, 
we locate the minimum number of appearances across all the transformation 
parameters ${\theta} \in param^T$ (i.e. $min\_count$ in Algorithm~\ref{alg:fault-loc}). 
This is to locate the worst-case degradation across all transformation parameters. 
The difference is then calculated between $min\_count$ and the number of appearances 
of the word in the original speech (i.e. $init\_count$). If the difference 
exceeds some user-defined threshold $\omega$, then \SMLT classifies the respective 
words as non robust w.r.t the group $GR$ and transformation $T$. 

We envision that practitioners can then review the data generated by fault localization 
(i.e. Algorithm~\ref{alg:fault-loc}) and target the non-robust words to further improve their 
ASR systems \revise{for speech from underrepresented groups~\cite{jain2018improved} and accommodate for speech variability~\cite{huang2001analysis}}. 
In {\bf RQ3}, we validate our fault localization method empirically and in 
{\bf RQ4}, we show how the proposed fault localization method can be used to highlight 
fairness violations.



\section{Datasets and Experimental Setup}
\label{sec:experimental-setup}

\smallskip \noindent
\textbf{ASR Systems under Test:}
We evaluate \SMLT on three commercial ASR systems from Google Cloud Platform (GCP), IBM Cloud, and Microsoft Azure. 
\revise{We use the standard models for GCP and Azure, and the {\em BroadbandModel} for IBM. }
In all three cases, the audio samples were identically encoded as .wav files using Linear 16 encoding.

\revise{In each of the following transformations, we vary a parameter, $\theta$. We call this the transformation parameter.}
\revise {
Some of the transformations have abbreviations within parentheses. 
Such abbreviations are used in later sections to refer to the respective transformations.}

\smallskip \noindent
\textbf{Amplitude Scaling (Amp)}: For amplitude scaling, we scale 
the audio sequence by a constant by multiplying
each individual audio sample by $\theta$.
%

\smallskip \noindent
\textbf{Clipping:} \revise{The audio samples are scaled such that their 
amplitude values are bound by [$-1$, $1$]. \SMLT then clips these samples such that the amplitude range is [$-\theta$, $\theta$]. 
These clipped samples are then rescaled and encoded.}


\smallskip \noindent
\textbf{Drop/Frame:} \revise{For Drop, 
\SMLT divides the audio into 20ms chunks. 
$\theta$\% of these chunks are then randomly discarded (amplitude set 
to zero) from the audio. For Frame, \SMLT divides the audio into $
\theta$ms chunks and 10\% of these chunks are then randomly discarded. No two adjacent chunks are discarded.}




\smallskip \noindent
\textbf{High Pass (HP)/ Low Pass (LP) Filter:} 
\revise{Here we apply a butterworth~\cite{butterworth} filter of order two to the entire audio file with $\theta$ determining the cut-off frequency.}




\smallskip \noindent
\textbf{Noise Addition (Noise):} \revise{
$\theta$ represents signal to noise 
(SNR) ratio~\cite{snr-def} of the transformed audio signal. 
A lower $\theta$ means higher noise in the transformed audio.}



\smallskip \noindent
\textbf{Frequency Scaling (Scale):} \revise{In this case, 
$\theta$ is the sampling frequency. The lower the value of $\theta$, the slower the audio. 
In this transformation, the audio is slowed down 
$\theta$ times.}


\revise{\Cref{tab:transList} lists all the different values used for 
$\theta$. An additional parameter ($\theta = 2.0$) is used for 
{\em Amp}.}

\begin{table}[t]
\caption{Transformations Used}
\centering
	{\scriptsize
    \begin{tabular}{l p{1cm}p{1cm}p{1cm}p{1cm}p{1cm}p{1cm}}
        \toprule
        \multicolumn{1}{l}{Transformation Type}  & \multicolumn{5}{c}{$\theta$ Used}  \\ \midrule
        \multicolumn{1}{l}{}  & \multicolumn{5}{c}{Least Destructive $\longrightarrow$ Most Destructive} \\  \cmidrule{2-6}
        Amplitude & 0.5 & 0.4 & 0.3 & 0.2 & 0.1 \\
        Clipping & 0.05 & 0.04 & 0.03 & 0.02 & 0.01 \\
        Drop & 5 & 10 & 15 & 20 & 25 \\
		Frame & 10 & 20 & 30 & 40 & 50 \\
		HP & 500 & 600 & 700 & 800 & 900 \\
		LP & 900 & 800 & 700 & 600 & 500 \\
		Noise & 10 & 8 & 6 & 4 & 2 \\
		Scale & 0.9 & 0.8 & 0.7 & 0.6 & 0.5 \\
		\bottomrule

    \end{tabular}}
    \label{tab:transList}
\end{table}

\begin{table}[t]
\caption{Datasets Used}
\centering
	{\scriptsize
    \begin{tabular}{l r rrr}
        \toprule 
        Dataset & \makecell{Duration(s)} & \#Clips & \makecell{\#Distinct\\Speakers} 
         \\ \midrule
        Accents  & 25-35 & 28 & 28  \\
        RAVDESS  & 3 & 32 & 8  \\
		Midlands  & 3-5 & 4 & 4 \\
		Nigerian English & 4-6 & 4 & 4 \\
		\bottomrule
    \end{tabular}}
    \label{tab:dataList}
\end{table}

\smallskip \noindent
\textbf{Datasets: }
\revise{We use the Speech Accent 
Archive (Accents)~\cite{accents}, the Ryerson Audio-Visual Database of Emotional Speech and Song (RAVDESS)~\cite{ravdess}, Multi speaker Corpora of the English Accents in the British Isles (Midlands)~\cite{uk-midlands-dataset}, and a Nigerian English speech dataset~\cite{nigerian-english} to evaluate \SMLT taking care to ensure male and female speakers are equally represented. \Cref{tab:dataList} provides additional details about the setup.}

\section{Results}
\label{sec:results}
\smallskip
In this section, we discuss our evaluation of \SMLT in detail. In particular, 
we structure our evaluation in the form of four research questions 
({\bf RQ1} to {\bf RQ4}). The analysis of these research questions appears in the 
following sections.

%
%
%
\smallskip \noindent
\textbf{\revise{RQ1: What is \SMLT's efficacy?}} 

We structure the analysis of this research question into three sections, 
each corresponding to a dataset we have used in our analysis. 
\revise{All of the relevant data is presented in \Cref{tab:smlt-errors}.}
We first analyse the number of errors (used interchangeably with fairness
violations) for each case. Subsequently, we 
analyse the sensitivity of the errors with respect to the values of 
$\tau$ ($\tau \in $ \{0.01, 0.05, 0.1, 0.15\}). 
\revise{Detecting violations of fairness is regulated by parameter $\tau$. Lower values of $\tau$ imply that the degradation of word error rates between two groups should be similar, and conversely higher values of $\tau$ allow for the difference in degradation of word error rates to be more severe between two groups.}
Next, we analyse the sensitivity of the pairs of the ASR systems under test. 
Concretely, we analyse the errors found in the  Microsoft Azure and IBM Watson 
{\em (MS\_IBM)}, Google Cloud and IBM Watson {\em (IBM\_GCP)}, and Microsoft 
Azure and Google Cloud {\em (MS\_GCP)} pairs. Finally, we analyse the sensitivity 
of the \SMLT test generation with respect to the eight different types of transformations 
implemented (see \Cref{fig:soundwave-transform}).

It is important to note that that we excluded the two 
most destructive Scale transformations. This is because the word error rate 
for these transformations is 0.89 on average out of 1. This degradation may 
be attributed to the transformation itself rather than the ASR. 
To avoid such cases, we exclude these transformations from this 
research question.

\begin{table*}[t]
\caption{Errors Discovered by \SMLT}
\centering
	\resizebox{\linewidth}{!}{
    \begin{tabular}{lr rrrrrrr rrr rr}
        \toprule
        \multicolumn{1}{l}{}  & \multicolumn{7}{c} {Accents} & \multicolumn{2}{c} {RAVDESS} & \multicolumn{2}{c} {\makecell{Nigerian/Midlands\\English}} \\ \cmidrule(lr){2-8} \cmidrule(lr){9-10} \cmidrule(lr){11-12}
        \multicolumn{1}{l}{}  & English & Ganda & French & Gujarati & 
        Indonesian &	Korean & 	Russian &  Male & Female & Midlands & Nigerian \\ \midrule
        
        \textbf{Total Errors} & 312 &	844 &	413 &
        406 &	311 & 1086	& 853 & 28 & 176 & 93 & 239\\
        \textbf{$\tau$ Sensitivity}  \\
         \multicolumn{1}{r}{\textit{0.01}} & 168 & 381 & 267 & 232 & 178 & 499 & 354 & 12 & 92  & 36 & 75 \\
		 \multicolumn{1}{r}{\textit{0.05}} & 75 & 245 & 99 & 101 & 85 & 340 & 227 & 8 & 53 & 26 & 65 \\
		 \multicolumn{1}{r}{\textit{0.10}} & 43 & 145 & 39 & 49 & 34 & 172 & 161 & 5 & 21 & 17 & 55\\ 
		 \multicolumn{1}{r}{\textit{0.15}} & 26 & 73 & 8 & 24 & 14 & 75 & 111 & 3 & 10 & 14 & 44\\ 
		 \textbf{ASR Sensitivity} \\
		 \multicolumn{1}{r}{\textit{MS IBM}} & 36 & 369 & 128 & 126 & 64 & 388 & 303 & 10 & 57 & 30 & 86 \\
\multicolumn{1}{r}{\textit{GCP IBM}} & 131 & 325 & 123 & 147 & 98 & 342 & 361& 9 & 64 & 31 & 96\\ 
\multicolumn{1}{r}{\textit{MS GCP}} & 145 & 150 & 162 & 133 & 149 & 356 & 189 & 9 & 55 & 32 & 57\\
		\textbf{Transition Sensitivty} \\
		\multicolumn{1}{r}{\textit{Clipping}} & 4 & 81 & 38 & 159 & 72 & 182 & 237 & 0 & 24 & 50 & 3\\
		\multicolumn{1}{r}{\textit{Drop}} & 8 & 113 & 33 & 29 & 40 & 184 & 45 & 0 & 21 & 4 & 33  \\
		\multicolumn{1}{r}{\textit{Frame}} & 14 & 106 & 61 & 25 & 36 & 170 & 26 & 1 & 13 & 13 & 19\\
		\multicolumn{1}{r}{\textit{Noise}} & 5 & 128 & 54 & 86 & 22 & 217 & 213& 0 & 24 & 5 & 43 \\
		\multicolumn{1}{r}{\textit{LP}} & 39 & 158 & 108 & 57 & 14 & 110 & 208 & 0 & 45 & 4 & 34  \\
		\multicolumn{1}{r}{\textit{Amplitude}} & 81 & 19 & 44 & 33 & 14 & 40 & 26 & 0 & 27 & 8 & 40\\
		\multicolumn{1}{r}{\textit{HP}} & 114 & 168 & 29 & 9 & 61 & 87 & 57 & 9 & 20 & 1 & 51  \\
		\multicolumn{1}{r}{\textit{Scale}} & 47 & 71 & 46 & 8 & 52 & 96 & 41 & 18 & 2 & 8 & 16 \\
        \bottomrule
    \end{tabular}}
    \label{tab:smlt-errors}
\end{table*}


\smallskip \noindent
\textbf{Accents Dataset:}
Native English speakers and Indonesian speakers have the lowest 
number of errors. 
%
On average, speech from non-native English speakers 
generates {\em 109\% more errors} in comparison to speech from native 
English speakers. 
%
%
For the two smallest values of $\tau$, speech from the native 
English speakers \revise{shows the {\em least number} of fairness violations}.
%
Speech from native English speakers has the lowest, second lowest and third 
lowest errors for the pairs of ASRs, {\em(MS\_IBM)}, {\em(MS\_GCP)} and 
{\em(IBM\_GCP)} respectively. 
%
Speech from native English speakers has the lowest errors for the 
clipping, two types of frame drops and noise transformations and the second 
lowest errors for the low-pass filter transformation.
The remaining transformations, namely amplitude, high-pass filter and 
scaling induce a comparable number of errors from native and 
non-native English speakers.

\begin{result}
Speech from non-native English speakers generally 
\revise{exhibits more fairness violations in} 
comparison to speech from native English speakers.
\end{result}
%

%

%

\smallskip \noindent
\textbf{RAVDESS Dataset:}
Speech from male speakers has significantly 
lower errors than speech from female speakers.
On average, speech from female speakers 
generates {\em 528.57\% more errors} in comparison to speech from male speakers.
%
%
%
\revise{
Speech from male speakers \revise{shows significantly fewer fairness violations} 
for all values of $\tau$, and for all ASR pairs tested.}
\revise{
Clipping, both types of frame drops, noise, low-pass and
amplitude induce significantly fewer errors on speech from male
speakers.}
%
However, speech from both groups have comparable number of errors when
subject to high-pass and scale transformations. 

\begin{result}
Speech from female speakers \revise{has significantly higher fairness violations} in 
comparison to speech from male speakers.
\end{result}

\smallskip \noindent
\textbf{Midlands/Nigeria Dataset:}
Speech from UK Midlands English (ME)
speakers has significantly lower errors 
than speech from Nigerian English (NE) speakers.
On average, speech from NE speakers
generates {\em 156.9\% more errors} in comparison to speech from 
ME speakers.
\revise{
Speech from ME speakers has significantly fewer fairness errors for all values of $\tau$, 
and for all ASR pairs tested.}
%
%
For the transformations scale, drop, noise, amplitude, low pass and high pass 
filters, the speech from ME speakers has significantly 
fewer error than speech from NE speakers.
For the transformations, clipping and frame, we find that speech from both 
groups have similar number of errors.

\begin{result}
Speech from Nigerian English speakers
\revise{has significantly more fairness errors} in 
comparison to speech from UK Midlands speakers.
\end{result}

\smallskip \noindent
\textbf{RQ2: What are the effects of transformations on 
comprehensibility?}

To better understand the effects of the transformations (\textit{see 
\Cref{fig:soundwave-transform}}) on the 
comprehensibility of the speech we conducted a user study. 
\revise{Speech of one {\em female native English} speaker from the 
Accents~\cite{accents} dataset was used. Survey participants 
were presented with the 
original audio file along with a set of transformed speech files in 
order of increasing intensity. All the transformations ({\em see 
\Cref{fig:soundwave-transform}}) and transformation parameters ({\em see 
\Cref{tab:transList}}) were used.}
We asked 200 survey participants (\revise{sourced through Amazon mTurk}) 
the following question:
\begin{result}
{\fontfamily{qcr}\selectfont
How comprehensible is (transformed) Speech with respect to the 
Original speech?}
\end{result}

\begin{wrapfigure}{l}{0.44\textwidth}
\vspace{-0.2in}
\begin{center}
\includegraphics[scale=0.37]{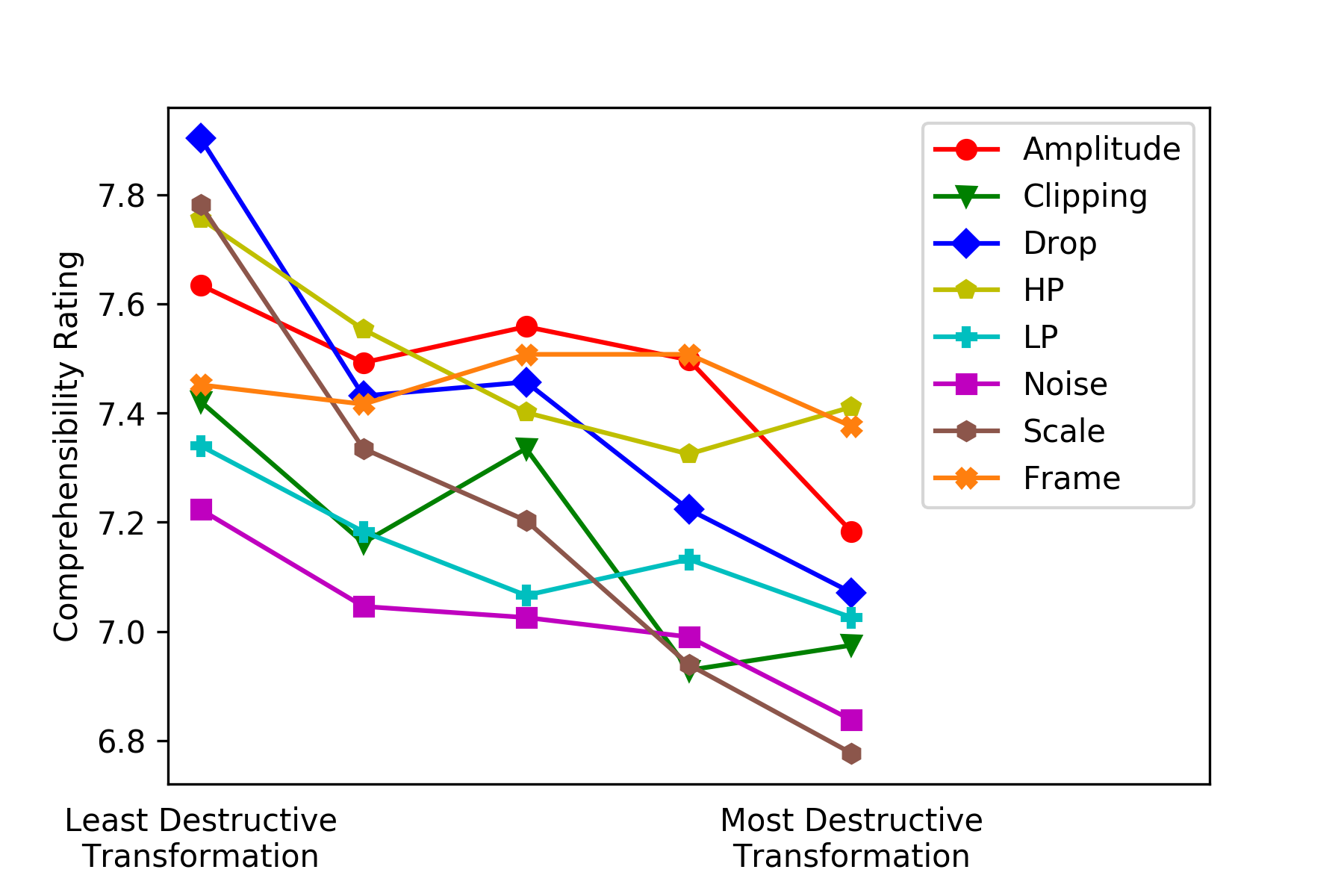}
\end{center}
\vspace{-0.2in}
\caption{Average Transformation Comprehensibility Ratings}
\label{fig:user-study-results}
\vspace{-0.2in}
\end{wrapfigure} 

The rating of one (1) is {\em Not Comprehensible at all} and the rating 
of ten (10) is {\em Just as Comprehensible as the Original}.


Unsurprisingly, as seen in \Cref{fig:user-study-results}, increasing 
the intensities of the transformation had a generally detrimental effect on
the comprehensibility of the speech. But none of the transformations 
majorly affect the comprehensibility of the speech. All of the 
transformations had a \revise{average} comprehensibility rating above 6.75 and 82.9\%
of the transformations had a comprehensibility rating above 7.


The average degradation in comprehensibility for the least 
destructive parameter across all transformations was 24.36\%. Noise
was the most destructive at 27.75\% and drop was the least
destructive (20.96\%). 

The average degradation in comprehensibility for the most 
destructive parameter across all transformations was 29.18\%. In this 
case, scaling 
was the most destructive at 32.23\% whereas drop was the least
destructive with 25.88\%. 

Additionally, for each transformation, we analyse the percentage 
drop of comprehensibility between the least and the most 
destructive transformation parameters. The average drop is 4.82\% 
across all transformations. The scaling and drop transformations 
show high relative percentage drops of 10.05\% and 8.32\% respectively. 
Amplitude, clipping, noise, high-pass and low-pass filters show closer 
to average drops between 3.1\% and 4.5\%. Frame, on the other hand, 
shows very low relative drops at 0.76\%.

\begin{result}
All the transformations, though destructive, are comprehensible by humans. 
\end{result}

For safety critical applications, we recommend that future work test the 
whole gamut of transformations. For other use cases, practitioners may choose
the transformations that satisfy their needs. 
To aid this, \SMLT allows the users to choose the comprehensibility 
threshold of the transformations. As seen in \Cref{tab:rq2-errors-compre}, 
our conclusion holds even if we choose the transformations with higher comprehensibility 
threshold (7.2). In particular, we observe that speech from native English speakers, 
male and UK Midlands Speakers generally exhibit lower errors. 
The detailed sensitivity analysis for the errors is seen in 
\Cref{fig:accents-errors-comp-thresh-7-2}, 
\Cref{fig:ravdess-errors-comp-thresh-7-2} and
\Cref{fig:nigeria-midlands-errors-comp-thresh-7-2} in the appendix. 
Additional user study details are seen in \Cref{sec:user-study-details}.


\begin{table*}[t]
\caption{Fairness errors where the transformations have a comprehensibility rating of at least 7.2}
\centering
	\resizebox{\linewidth}{!}{
    \begin{tabular}{lr rrrrrrr rrr rr}
        \toprule
        \multicolumn{1}{l}{}  & \multicolumn{7}{c} {Accents} & \multicolumn{2}{c} {RAVDESS} & \multicolumn{2}{c} {\makecell{Nigerian/Midlands English}} \\ \cmidrule(lr){2-8} \cmidrule(lr){9-10} \cmidrule(lr){11-12}
        \multicolumn{1}{l}{}  & English & Ganda & French & Gujarati & 
        Indonesian &	Korean & 	Russian &  Male & Female & Midlands & Nigerian \\ \midrule
        
        \textbf{\makecell{Total Errors}} & 246 &	509 &	240 &
        166 &	225 & 687	& 329 & 28 & 88 & 55 & 161\\
        \bottomrule
    \end{tabular}}
    \label{tab:rq2-errors-compre}
\end{table*}

\begin{table}[t]
\caption{Grammar-generated sentence examples}
\centering

\resizebox{\linewidth}{!}{\scriptsize
	\begin{tabular}{lrp{4cm}p{4cm}p{4cm}}
	\toprule
	 \multicolumn{2}{l}{ASR}  & \multicolumn{1}{l}{Microsoft} 
	 &\multicolumn{1}{l}{Google Cloud} 
	 & \multicolumn{1}{l}{IBM Watson} \\ \midrule
	 & \textit{Robust} & Ashley likes \textbf{fresh} smoothies & Karen loves \textbf{plastic} straws & William detests \textbf{plastic} cups\\ 
     & & Paul adores \textbf{spoons} of cinnamon & Donald hates 
     \textbf{big} decisions & Steven detests \textbf{big} flags  \\ \\
     & \textit{Non-robust}  &  Ashley detests \textbf{thick} smoothies &
     John loves \textbf{spoons} of cinnamon &
     Betty likes \textbf{scoops} of ice cream \\ 
     & & Ryan likes \textbf{slabs} of cake  & Robert loves \textbf{bags} 
     of concrete & Amanda is fond of \textbf{things} like groceries   \\ 
	 \bottomrule
	\end{tabular}

}
\label{tab:gram-sentences}
\end{table}

\smallskip \noindent
\textbf{RQ3: Are the outputs produced by \SMLT fault localiser 
valid?}

To study the validity of the outputs of the fault localiser, 
we study the number of errors for the predicted robust and 
non-robust words. 
We do this by generating speech containing the 
predicted robust and non-robust words for each ASR tested. 
We choose an $\omega$ of three, three and two for GCP, MS Azure and IBM
respectively to choose the non-robust words ({\em see 
\Cref{alg:fault-loc}}). We choose the robust 
words from the set of words that do not show any errors in the presence
of noise ($\mathit{count\_diff}$ = 0 in {\em 
\Cref{alg:fault-loc}}) for these specific ASR systems.
Specifically, we test whether the robust and 
non-robust words identified by the fault localiser in the 
Accents dataset are robust in the presence of noise.
Our goal is to show that if noise is added to speech 
containing these non-robust words, the ASR will be less likely to recognise 
them. Vice-versa, if noise is added to the predicted robust-words they are 
less likely to be affected. 

To generate the speech from the output we generate sentences containing
the robust and non-robust words predicted by the fault localiser 
for each ASR using a grammar and then 
use a text-to-speech (TTS) service to generate speech. The actual randomly 
selected robust and non-robust words (in bold) and the examples
of the sentences generated by the grammar can be seen in 
\Cref{tab:gram-sentences}. The grammars themselves can be seen in 
\Cref{fig:rq3-grammars}.
We use the Google TTS for MS Azure and we use the 
Microsoft Azure TTS for GCP and IBM to generate the speech.

To evaluate the generality of outputs of the fault 
localisation technique, we use the speech produced by the TTS  
and then add noise to that speech. This speech is used to generate a 
transcript from the ASR and the transcript is used to evaluate how 
many of the predicted robust and non-robust words are incorrect in the 
transcript. We add the most noise possible to the TTS speech in our \SMLT 
framework. Specifically, the signal to noise (SNR) ratio is 2. We use 
the TTS generated speech for 50 sentences for each of the robust and 
non-robust cases. Each sentence has either a robust or a non-robust 
word.

The results of the experiments are seen in \Cref{tab:explanation-generality}.
In the transcript of the speech with noise added at SNR 2, robust words 
show {\em zero} error for the predicted robust words for Microsoft and 
Google Cloud and 21 errors for IBM. The non-robust words
on the other hand had 23, 15 and 30 errors. Thus, the predicted
non-robust words have a higher propensity for errors than the 
robust words. 

\begin{result}
	The outputs of the fault localisation techniques are general and valid.  
\end{result}

\begin{table}
	\begin{minipage}[t]{0.4\textwidth}
	\caption{Transcript Errors}
	\centering
	{\scriptsize
    \begin{tabular}{lrrrr}
        \toprule
        \multicolumn{2}{l}{ASR}  & \makecell{Transcript\\Errors} \\ \midrule
        \multicolumn{2}{l}{Microsoft (MS)} & \\
        & \textit{Robust} & 0 \\ 
        & \textit{Non-Robust} & 23  \\ 
        \cmidrule{1-3}
        \multicolumn{2}{l}{Google Cloud (GCP)} & \\
         & \textit{Robust} & 0\\ 
        & \textit{Non-Robust} & 15\\ 
        \cmidrule{1-3}
        \multicolumn{2}{l}{IBM Watson (IBM)} & \\
        & \textit{Robust} & 21  \\ 
        & \textit{Non-Robust} & 30 \\ 
        \bottomrule
    \end{tabular}}
    \label{tab:explanation-generality}
	\end{minipage}
	\centering
	\begin{minipage}[t]{0.55\textwidth}
	\caption{Grammarly Scores}
	\centering
	{\scriptsize
	\centering
    \begin{tabular}{lrrrr}
        \toprule
        \multicolumn{2}{l}{ASR}  & \makecell{Overall\\Score} & 
        \makecell{Correctness} & \makecell{Clarity} \\ \midrule
        \multicolumn{2}{l}{Microsoft (MS)} & \\
        & \textit{Robust} & 99 & \multirow{8}{*}{\makecell{Looking\\Good}} &  
        \multirow{8}{*}{\makecell{Very\\Clear}}\\ 
        & \textit{Non-Robust} & 99 & &  \\ 
        \cmidrule{1-3}
        \multicolumn{2}{l}{Google Cloud (GCP)} & \\
         & \textit{Robust} & 100 &  & \\ 
        & \textit{Non-Robust} & 99 & & \\ 
        \cmidrule{1-3}
        \multicolumn{2}{l}{IBM Watson (IBM)} & \\
        & \textit{Robust} & 100 & &  \\ 
        & \textit{Non-Robust} & 96 & & \\ 
        \bottomrule
    \end{tabular}}
    \label{tab:gram-eval}
	\end{minipage}
\end{table}

\smallskip \noindent
\textbf{Note on grammar validity:}
Since the grammars used by us to validate the explanations of 
\SMLT are handcrafted, they may be prone to errors. To verify these 
hand crafted grammars, we use 100 sentences produced by each grammar 
and use the online tool Grammarly~\cite{grammarly} to investigate the 
semantic and syntactic correctness of the sentences and the clarity. 
The sentences generated by the grammars have a high overall average 
score of 98.33 out of 100, with the lowest being 96 
(\textit{see \Cref{tab:gram-eval}}). On the correctness and clarity 
measure, all the sentences generated by the grammars score 
{\em Looking Good} and {\em Very Clear}. 
%


\smallskip \noindent
\textbf{RQ4: Can the fault localiser be used to 
highlight unfairness?}

\begin{table*}[t]
\caption{Average words mispredictions in the Accents dataset using the \SMLT localisation techniques}
\centering
	{ \scriptsize
    \begin{tabular}{lr rrrrrrr rrr rr}
        \toprule
        \multicolumn{1}{l}{}  & \multicolumn{7}{c} {Accents} \\ \cmidrule(lr){2-8}
        \multicolumn{1}{l}{}  & English & Ganda & French & Gujarati & 
        Indonesian &	Korean & 	Russian 
        \\ \midrule
        
        \textbf{\makecell{ASR Sensitivity}} \\
        \multicolumn{1}{r}{\textit{GCP}} & 1.21 & 1.51 & 1.21 & 1.17 & 1.07 & 1.55 & 1.64\\
        \multicolumn{1}{r}{\textit{IBM}} & 1.03 & 1.94 & 1.38 & 1.35 & 1.48 & 1.92 & 1.70 \\
        \multicolumn{1}{r}{\textit{MS Azure}} & 0.47 & 0.66 & 0.40 & 0.48 & 0.36 & 0.87 & 0.63 \\
        \textbf{\makecell{Transition Sensitivity}} \\
        
        \multicolumn{1}{r}{\textit{Clipping}} & 2.00 & 2.53 & 2.12 & 2.60 & 2.29 & 2.81 & 3.13 \\
		\multicolumn{1}{r}{\textit{Drop}} & 0.30 & 1.02 & 0.52 & 0.54 & 0.57 & 1.15 & 0.74 \\
		\multicolumn{1}{r}{\textit{Frame}} & 0.38 & 0.89 & 0.68 & 0.56 & 0.51 & 1.19 & 0.65 \\
		\multicolumn{1}{r}{\textit{Noise}} & 0.57 & 1.60 & 0.85 & 1.27 & 0.71 & 1.74 & 1.54 \\
		\multicolumn{1}{r}{\textit{LP}} & 1.72 & 2.22 & 1.90 & 1.79 & 1.58 & 1.98 & 2.13 \\
		\multicolumn{1}{r}{\textit{Amplitude}} & 0.17 & 0.15 & 0.11 & 0.12 & 0.06 & 0.20 & 0.16 \\
		\multicolumn{1}{r}{\textit{HP}} & 0.74 & 0.75 & 0.38 & 0.22 & 0.49 & 0.64 & 0.76 \\
		\multicolumn{1}{r}{\textit{Scale}} & 1.38 & 1.79 & 1.42 & 0.90 & 1.54 & 1.89 & 1.45 \\
        
        \bottomrule
    \end{tabular}}
    \label{tab:localiser-accents-errors}
\end{table*}

The goal of this RQ is to investigate if the output of \Cref{alg:fault-loc}
can call attention to bias between different groups. Specifically, we 
evaluate if some groups show fewer faults, on average than others. 
To this end, we use the fault localisation algorithm 
(\textit{\Cref{alg:fault-loc}}) on the accents dataset and 
record the number of words incorrect in the transcript, on average for
each group of the accents dataset. This is done for each ASR under test. 
It is also important to note that 
this technique uses no ground truth data and requires no manual input. 
This technique is designed to work with just the speech data and metadata
(groups).

\Cref{tab:localiser-accents-errors} shows the average word drops across 
all transformations for the accents dataset for each ASR under test. 
Speech from native-English speakers shows the lowest average word drops for
the IBM Watson ASR and the third lowest for GCP and MS Azure ASRs.
%
We also investigate the average word drops for each transformation 
in \SMLT averaged across all ASRs. 
Speech from native English speakers has the lowest average word drops for the 
Clipping, two types of frame drops and noise transformations and the second 
lowest errors for the low-pass filter transformation. 
(\textit{see \Cref{tab:localiser-accents-errors}}). 
%
For the rest of the transformations, namely amplitude, high-pass filter and 
scaling, we find that both speech from non-native English speakers and speech from native 
English speakers have comparable average word drops (\textit{see 
\Cref{tab:localiser-accents-errors}}). 
This result is consistent with results
seen in RQ1.

\begin{result}
The technique seen in \Cref{alg:fault-loc} can be used to highlight bias 
in speech and the results are consistent with RQ1.	
\end{result}

\section{Threats to Validity}
\label{sec:threats}

\smallskip \noindent
\textbf{User Study:} In conducting the study, two assumptions were made. 
Firstly, we assume that the degree to which comprehensibility changes when 
subject to transformations is independent of the characteristics of the 
speaker’s voice. Secondly, we assume that the speech is reflective of the 
broader English language. 
In future work a larger scale user study could 
be performed to verify the results.

\smallskip \noindent
\textbf{ASR Baseline Accuracy:} 
\SMLT measures the degradation of the speech to characterise
the unfairness amongst groups and ASR systems. If the baseline error rate 
is very high, then the room for further degradation is very low. 
\revise{As a result, \SMLT expects 
ASR services to have a high baseline accuracy. To mitigate this threat, we
use state-of-the-art commercial ASR systems which have high baseline accuracies.}

\smallskip \noindent
\textbf{Completeness and Speech Data:}
\SMLT is incomplete, by design, in the discovery 
of fairness violations. \SMLT is limited by the speech data and the groups
of this speech data used to test these ASR systems. With new data and 
new groups, it is possible to discover more fairness violations. 
The practitioners need to provide data to discover these. In our 
view, this is a valid assumption because the developers of these 
systems have a large (and growing) corpus of such speech data. It is 
also important to note that \SMLT does not need the ground truth 
transcripts for this speech data and such speech data is easier to obtain. 

\smallskip \noindent
\textbf{Fault Localisation:}
To test \SMLT's fault localisation, we identify the robust and non-robust 
words in the speech and subsequently construct sentences (with the aid 
of a grammar). These sentences are then converted to speech using
a text-to-speech (TTS) software and the performance of the robust and non-
robust words is measured. In the future, we would like to repeat the same 
experiment with a fixed set of speakers, which allows us capture the 
peculiarities of speech in contrast to the usage of TTS software.




\section{Related Work}

In the past few years, there has been significant attention in testing 
ML systems \cite{deepXplore,deeptest,deepgauge,deepconcolic,deephunter,tensorfuzz,ogma,checklist,mettle,deepgini,robotDLTesting,calo2020simultaneously,sharma2021mlcheck,Guo2020AAT}. 
Some of these works target coverage-based 
testing~\cite{deeptest,deephunter,tensorfuzz,deepgauge} \revise{or
leverage property driven testing~\cite{sharma2021mlcheck}}, while others focus
effective testing in targeted domains e.g. text~\cite{ogma,checklist}. 
None of these works, however, are directly applicable for testing ASR systems. 
In contrast, the goal of \SMLT is to automatically discover 
violations of fairness in ASR systems without access to ground truth data. 

DeepCruiser~\cite{deepcruiser} uses metamorphic transformations
and performs coverage-guided fuzzing to discover transcription errors 
in ASR systems. Concurrently, CrossASR~\cite{crossASR} uses text to 
generate speech from a TTS engine and subsequently employs differential 
testing to find bugs in the ASR system.
In contrast to these systems, the goal of \SMLT is to automatically find
violations of fairness by measuring the degradation of transcription 
quality from the ASR when the speech is transformed. \SMLT compares 
this degradation across various groups of speakers and if the 
difference is substantial, \SMLT characterises this as a fairness
violation. 
Moreover, \SMLT neither requires access 
to manually labelled speech data nor does it require any white/grey box 
access to the ASR model.
%
%
Works on audio adversarial testing~\cite{automatedBasicRecognition}, 
\cite{fakeBob}, \cite{audioAdversarial}, \cite{imperceptibleASRadversarial}, 
\cite{foolingAsr} aims to find
an imperceptible perturbation that are specially crafted for an audio file. 
In contrast, \SMLT aims to find fairness violations. Additionally, \SMLT 
also proposes automatic fault 
localisation for ASR systems without using a ground truth transcript.



%
\revise{Unlike \SMLT, recent works on fairness testing have focused 
on credit rating~\cite{themis,aequitas,blackboxFainess,ignoranceAndPrejudice,sharma2019testing,sharma2020higher,sharma2020automatic,sharma2021mlcheck}, 
computer vision~\cite{denton2019image,buolamwini2018gender} or NLP systems~\cite{ijcai_nlp_testing,astraea}.
In the systems that deal with such data, it is possible to isolate certain
sensitive attributes (gender, age, nationality) and test for 
fairness based on these attributes. It is challenging to isolate such 
sensitive attributes in speech data, necessitating the need for a separate
fairness testing framework specifically for speech data.}


Frameworks such as LIME~\cite{lime}, SHAP~\cite{shap}, Anchor~\cite{anchor} and DeepCover~\cite{eccv2020_fault}
attempt to reason why a model generates a specific
output for a specific input.
In contrast to this, \SMLT's fault localisation algorithm 
identifies utterances spoken 
by a group which are likely to be not recognised by ASR systems in the 
presence of a destructive interference (such as noise).
\revise{Recent fault localization approaches either aim to  
highlight the neurons~\cite{fase2019_fault} or training code~\cite{icse2021_fault} 
that are responsible for a fault during inference. 
In contrast, \SMLT highlights words that are likely to be transcribed wrongly without 
having any access to the ground truth transcription and with only blackbox access to 
the ASR system.} 



\section{Conclusion}
\label{sec:conclusion}

In this work we introduce \SMLT, an automated fairness testing 
technique for ASR systems. To the best of our knowledge, 
we are the first work that explores considerations beyond error
rates for discovering fairness violations. We also show that the 
speech transformations used by \SMLT are largely comprehensible
through a user study. 
Additionally, \SMLT highlights words where a given ASR system exhibits
faults, and we show the validity of these explanations. These faults 
can also be used to identify unfairness in ASR systems. 

\SMLT is evaluated on three ASR systems and we use four distinct datasets. 
Our experiments reveal that speech from non-native English, female and Nigerian 
English speakers exhibit more errors, on average than speech from native English, 
male and UK Midlands speakers, respectively. We also validate the fault 
localization embodied in \SMLT by showing that the predicted non-robust 
words exhibit {\em 223.8\%} more errors than the predicted robust words 
across all ASRs.


We hope that \SMLT drives further work on systematic fairness testing of 
ASR systems. To aid future work, we make all our code and data publicly 
available: \textbf{https://github.com/sparkssss/AequeVox}
%

%
%
%
\bibliographystyle{splncs04}
\bibliography{ms}
 
\appendix
\newpage
\section{Additional Tables}

\begin{table}[h]

\caption{Average User Study Comprehensibility Scores}
\centering
	{\scriptsize
    \begin{tabular}{l p{1cm}p{1cm}p{1cm}p{1cm}p{1cm}p{1cm}}
        \toprule
        \multicolumn{1}{l}{Transformation}  & \multicolumn{5}{c}{Average Comprehensibility Score}  \\ \midrule
        \multicolumn{1}{l}{}  & \multicolumn{5}{c}{Least Destructive $\longrightarrow$ Most Destructive} \\  \cmidrule{2-6}
        Amplitude & 7.63 & 7.49 & 7.56 & 7.50 & 7.18 \\
        Clipping & 7.42 & 7.16 & 7.34 & 6.93 & 6.97 \\
        Drop & 7.90 & 7.43 & 7.46 & 7.22 & 7.07 \\
		Frame & 7.45 & 7.42 & 7.51 & 7.51 & 7.38 \\
		HP & 7.76 & 7.55 & 7.40 & 7.32 & 7.41 \\
		LP & 7.34 & 7.18 & 7.07 & 7.13 & 7.03 \\
		Noise & 7.22 & 7.05 & 7.03 & 6.99 & 6.84 \\
		Scale & 7.78 & 7.34 & 7.20 & 6.94 & 6.78 \\
		\bottomrule

    \end{tabular}}
    \label{tab:user-study-score}
    
\end{table}

\newpage
\section{Sound Transformations}
\label{sec:sound-transformations}
\smallskip \noindent
\textbf{Sound Wave:} To understand metamorphic transformations of sound, it 
is useful to understand the sinusoidal representation of sound. 
A sound wave of a single amplitude and frequency can be represented as follows: 
\begin{align}
	y(t) = A \sin(2\pi ft+\phi)
	\label{eqn:sinusoid}
\end{align}
where $A$ is the amplitude, the peak deviation of the function from zero, 
$f$ is the ordinary frequency i.e. the number of oscillations (cycles) 
that occur each second and $\phi$ is the phase which specifies (in radians)
where in its cycle the oscillation is at time $t = 0$.

It is known that any sound can be expressed as a sum of 
sinusoids~\cite{spectrum-sinusoids}. 
The transformations on sinusoidal wave can thus, be applied to any sound. 
Without losing generality and for simplicity we only show the transformations 
for a sound wave captured by a single sinusoidal wave. This is the wave of 
the form seen in \Cref{eqn:sinusoid}. To have a variable frequency, we set 
$f \propto \frac{t}{c}$ where $c > 1$ and $c \in \mathbb{R}$. This wave is 
seen in \Cref{fig:overview}~(a).

In the following, we describe the transformations used in our \SMLT technique. 

\smallskip \noindent
\textbf{Noise Addition:}
Noise robust ASR systems is a classic field of research and in the past 
thirty years there have been to the order of a hundred different techniques
to try and solve this problem~\cite{noise-robust}. 
Noise is also a natural phenomenon in daily life and we may not expect signals
used by ASR systems to be totally clean. 
As a result, one expects an ASR system to take noise into account and still
be effective in noisy environments. 

%
At each time step $t$ in the sound wave, a random variable $R \sim 
\mathbb{D}$, where $\mathbb{D}$ is some distribution, is added. 
As the range of $R$ increases, the noise increases and the signal
to noise ratio decreases. The metamorphic transformation of adding 
noise is seen in \Cref{fig:soundwave-transform}~(b). Concretely 
the transformed function $y^T(t)$ can be expressed as follows: 
\begin{align}
	y^T(t) = y(t) + R \qquad \forall t, \ R \sim \mathbb{D}
	\label{eqn:noise}
\end{align}


\smallskip \noindent
\textbf{Amplitude Modification:} A sound wave's amplitude relates to the 
changes in pressure. A sound is perceived as louder if the amplitude 
increases and softer if it decreases. We expect ASR systems to have minor 
degradations in performance, if any across groups of loud and soft speakers.  
To this end, as seen in \Cref{fig:soundwave-transform}~(c)., we
increase or decrease the amplitude of a sound wave as a metamorphic 
transformation. Concretely the transformed function $y^T(t)$ can be 
expressed as follows: 
\begin{align}
	y^T(t) = c*y(t) \qquad \forall t, c\in \mathbb{R}
	\label{eqn:amplitude}
\end{align}


\smallskip \noindent
\textbf{Frequency Scaling:} 
In this type of distortion, the frequency of the audio signal is scaled up
or down by some constant factor. We expect ASR systems to be largely 
robust to changes in frequency (slowing down or speeding up) in the
speech signal (see \Cref{fig:soundwave-transform}~(d)). To this end, we modify 
the frequency of a sound as a metamorphic transformation as follows: 
\begin{align}
	y^T(t) = y(c*t) \qquad \forall t, c\in \mathbb{R}
	\label{eqn:frequency}
\end{align}


\smallskip \noindent
\textbf{Amplitude Clipping:} Clipping is a form of distortion that limits 
the signal once a threshold is exceeded. For sound, once the wave exceeds
a certain amplitude, the sound wave is clipped. 
Clipping occurs when the sound signal exceeds the maximum dynamic range of 
an audio channel~\cite{modern-recording}. To simulate this,
we use clipping as a metamorphic transformation as follows (see \Cref{fig:soundwave-transform}~(e)): 
\begin{align}
    y^T(t)& = 
    \begin{cases}
       c, &  y(t) > c,\\
       y(t), & -c < y(t) < c, \\
       -c, &  y(t) < c,
    \end{cases} 
    \qquad \forall t,  c \in \mathbb{R}
    \label{eqn:amplitude-clipping}
\end{align}

\smallskip \noindent
\textbf{Frame Drop:} A common scenario with wireless communication is 
the dropping of information (frames or samples in technical
parlance~\cite{communications-dict})
due to interference with other signals. This usually happens when a 
signal is modified in a disruptive manner. A common example of this 
is a crosstalk on telephones. To simulate this effect as a metamorphic 
transformation for the ASR system, \SMLT randomly drops some frames and 
information to test for the robustness of the system. This metamorphic 
transformation is seen in \Cref{fig:soundwave-transform}~(f). Formally, 
the transformation is captured as follows: 
\begin{align}
    y^T(t)& = 
    \begin{cases}
       y(t), &  t \not\in \mathit{FD},\\
       0, &  t \in \mathit{FD},\\
    \end{cases} 
    \qquad \forall t
    \label{eqn:frame-drop}
\end{align}
where $\mathit{FD}$ is a set which contains the values of $t$ where the 
frames are dropped. The set $\mathit{FD}$ can be configured by the user, 
or randomly.
There are two considerations to be made when performing the  
transformation in \Cref{eqn:frame-drop}. The first is the total percentage 
of the signal to be dropped, $tot\_drop$. This means that out of the total 
length of the signal, the transformation drops $tot\_drop$\% of the signal. 
The second is $frame\_size$, which controls the size of continuous 
signal that is dropped. \SMLT considers both the aforementioned cases. 
Specifically, in one case \SMLT keeps $tot\_drop$ constant and varies the 
$frame\_size$, while in the other, we keep $frame\_size$ constant and vary 
the $tot\_drop$ percentage. 

\smallskip \noindent
\textbf{High/Low-Pass filters:} High-pass filters only let sounds with 
frequencies higher than a certain threshold pass, and conversely low-pass
filters only let sounds with frequencies lower than a certain threshold 
pass. These filters are commonly used in audio 
systems to direct frequencies of sound to certain types of speakers. This 
is because speakers are designed for certain types of frequencies and 
sound waves outside of those frequencies might damage these speakers. 
In our evaluation, to simulate the source of sound being from one of such 
speakers, we use these filters as a metamorphic transformation. The low-pass
filter transformation is seen in \Cref{fig:soundwave-transform}~(g) and the 
high pass filter transformation is seen in
\Cref{fig:soundwave-transform}~(h). The transformation equation 
for the high-pass filter is seen in \Cref{eqn:hp-filter} and 
correspondingly, for the low-pass filter is seen in \Cref{eqn:lp-filter}. 
$\Theta_{HP}$ and $\Theta_{LP}$ are the high pass and low pass filter
thresholds, respectively.
\begin{align}
    y^T(t)& = 
    \begin{cases}
       y(t), &  f > \Theta_{HP},\\
       0, &  f < \Theta_{HP},\\
    \end{cases} 
    \qquad \forall t
    \label{eqn:hp-filter}
\end{align}

\begin{align}
    y^T(t)& = 
    \begin{cases}
       y(t), &  f < \Theta_{LP},\\
       0, &  f > \Theta_{LP},\\
    \end{cases} 
    \qquad \forall t
    \label{eqn:lp-filter}
\end{align}

\newpage
\section{User Study Setup Details} 
\label{sec:user-study-details}
We conducted a user study using Amazon’s mTurk platform. In particular, 200 participants were presented with an audio file containing speech utterances by a {\em female native English speaker}. In addition, the audio clip contained nearly all the sounds in the English language to represent the full spectrum of the language, as found in Speech Accent Archive~\cite{accents}. Users were presented with the original audio file along with a set of transformed speech files in order of increasing intensity. For instance, in the case of the "Scale" transformation, participants were first presented with a file that was slightly slowed down and subsequent files were slowed down even further. Users then rated the comprehensibility of the speech files in comparison to the original audio file. The rating was on a 1 to 10 scale, where "10" refers to the case where the modified speech file was just as comprehensible as the original speech and "1" refers to the case where 
the modified speech was not comprehensible at all.  

Participants were required to rate the comprehensibility of the entire set of transformations under 
study i.e. Amplitude, Clipping, Drop, Frame, Highpass, Lowpass, Noise and Scale 
(see~\Cref{fig:soundwave-transform}). The average score of each transformation was used to determine 
the comprehensibility score. In general, we see that 
the comprehensibility of the speech tends to go down as the intensity of the transformation increases, 
as observed in~\Cref{fig:user-study-results}. We present a comprehensive analysis of the user study results in {\bf RQ2}.

\newpage
\section{Additional Figures}
\label{sec:additional-figs} 

\begin{figure}[h]
\begin{center}
\begin{tabular}{c}
\includegraphics[scale=0.7]{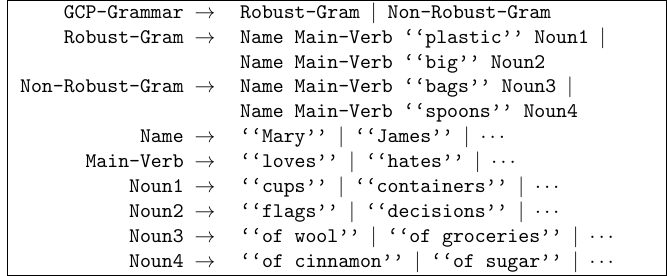} \\
{\bf (a)} \\ \\
\includegraphics[scale=0.7]{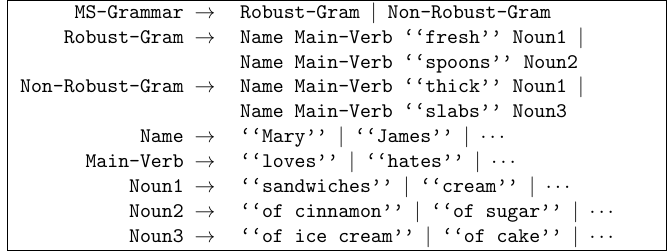} \\
{\bf (b)} \\ \\
\includegraphics[scale=0.7]{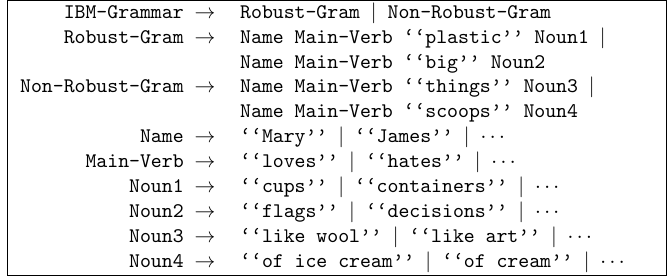} \\
{\bf (c)}\\
\end{tabular}
\caption{Grammars used by \SMLT to verify the generality of the Fault localiser predictions}
\end{center}
\label{fig:rq3-grammars}
\end{figure}

\begin{figure*}[h]
\begin{center}
{
\begin{tabular}{cc}
\includegraphics[scale=0.25]{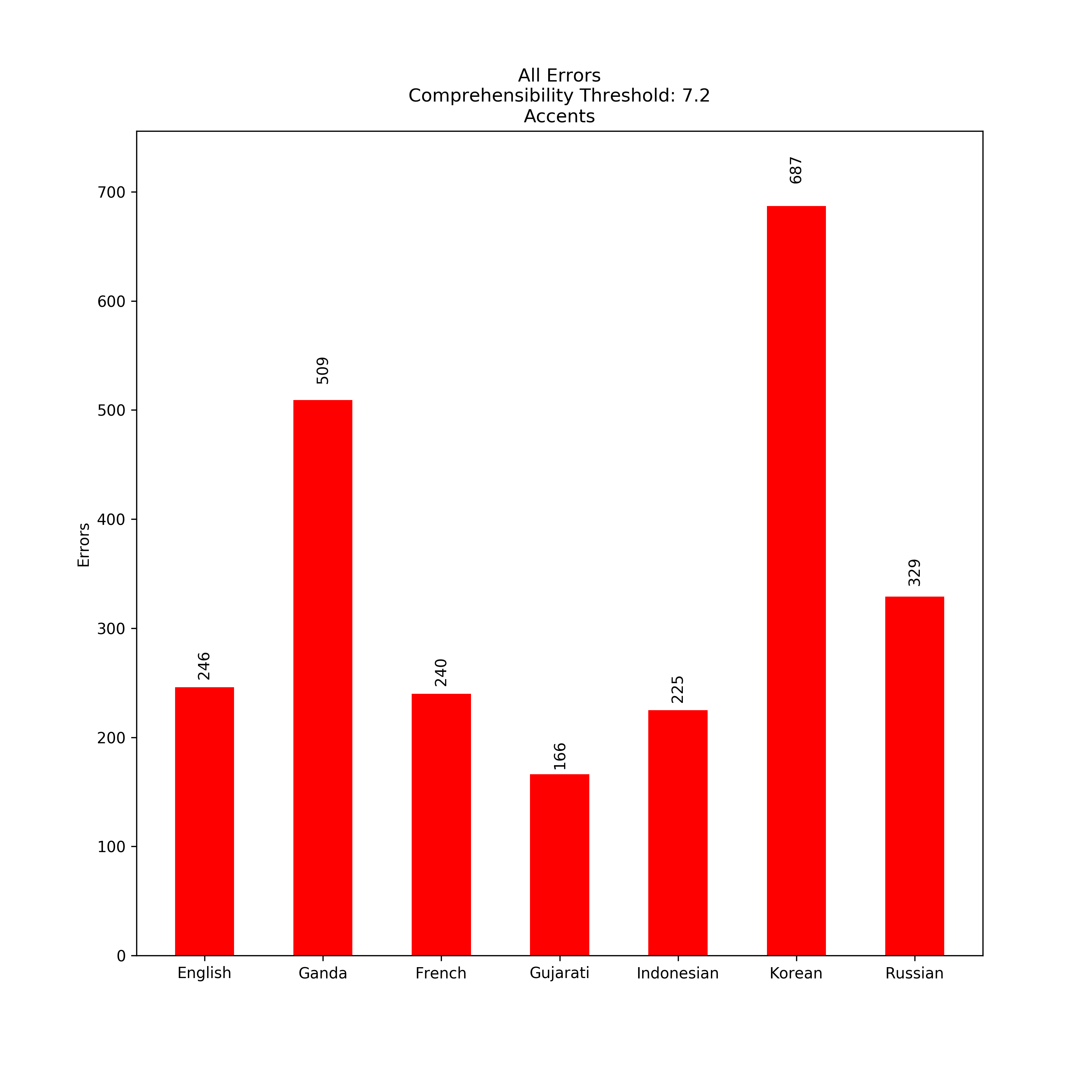} &
\includegraphics[scale=0.25]{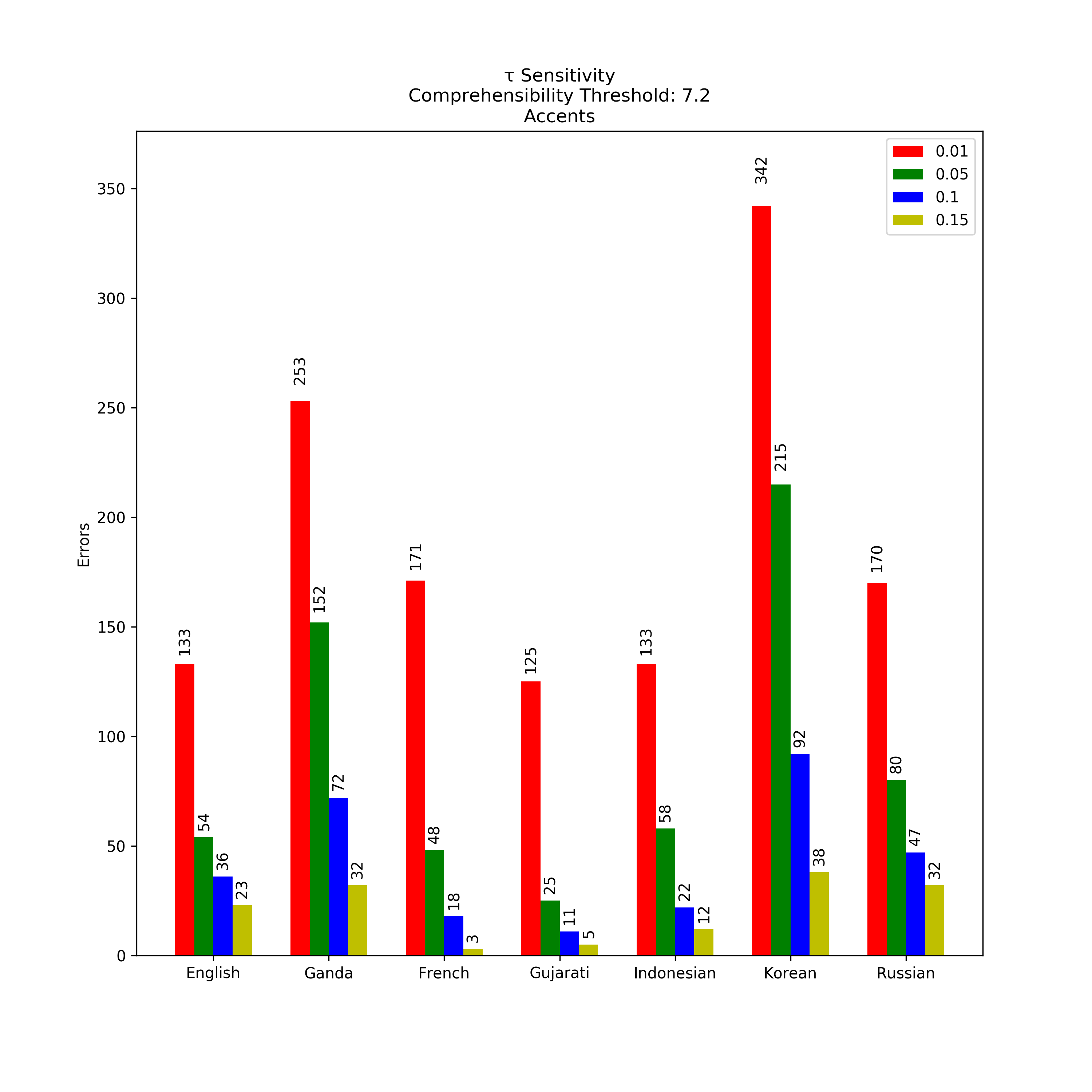} \\
{\bf (a)} & {\bf (b)}\\
\includegraphics[scale=0.25]{{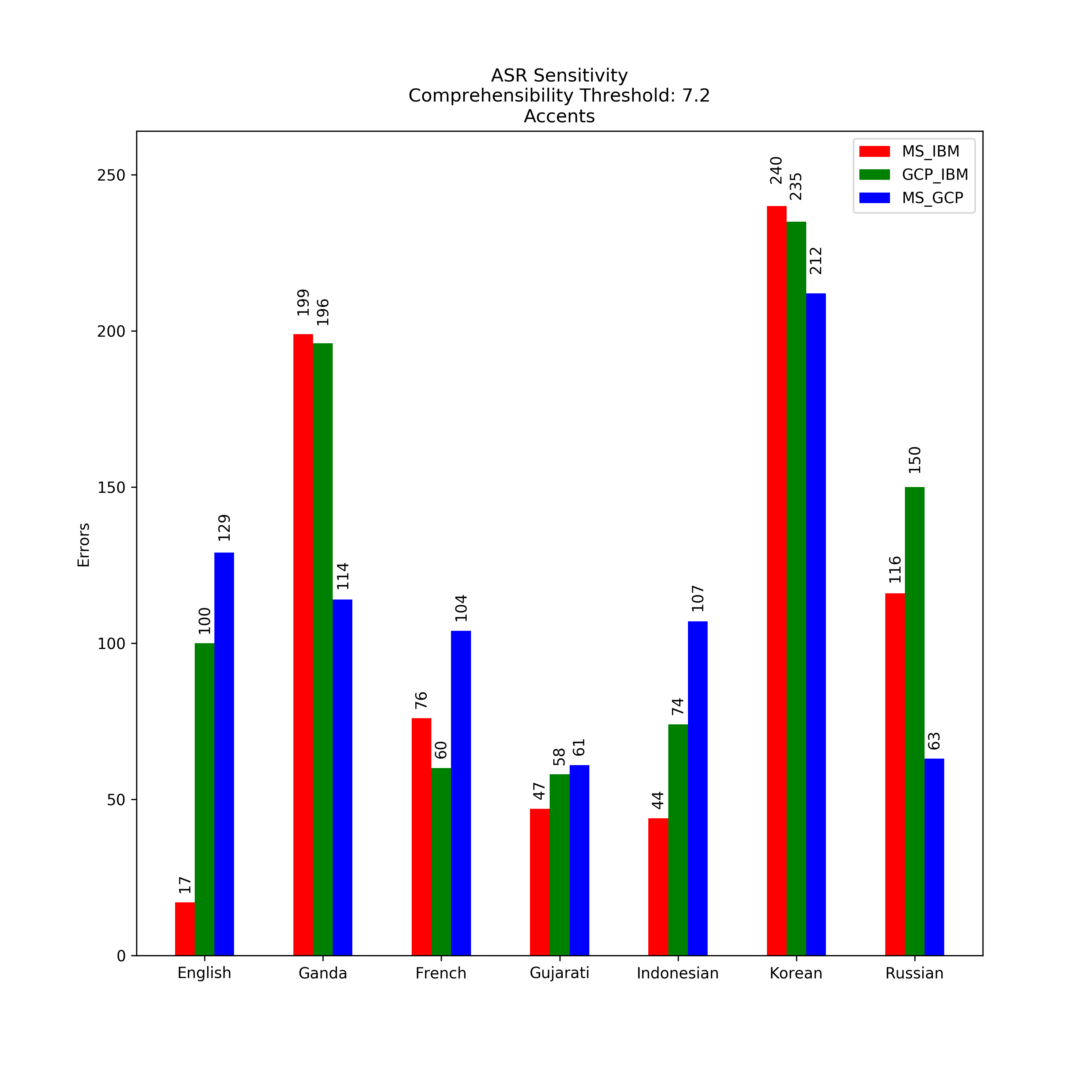}} &
\includegraphics[scale=0.25]{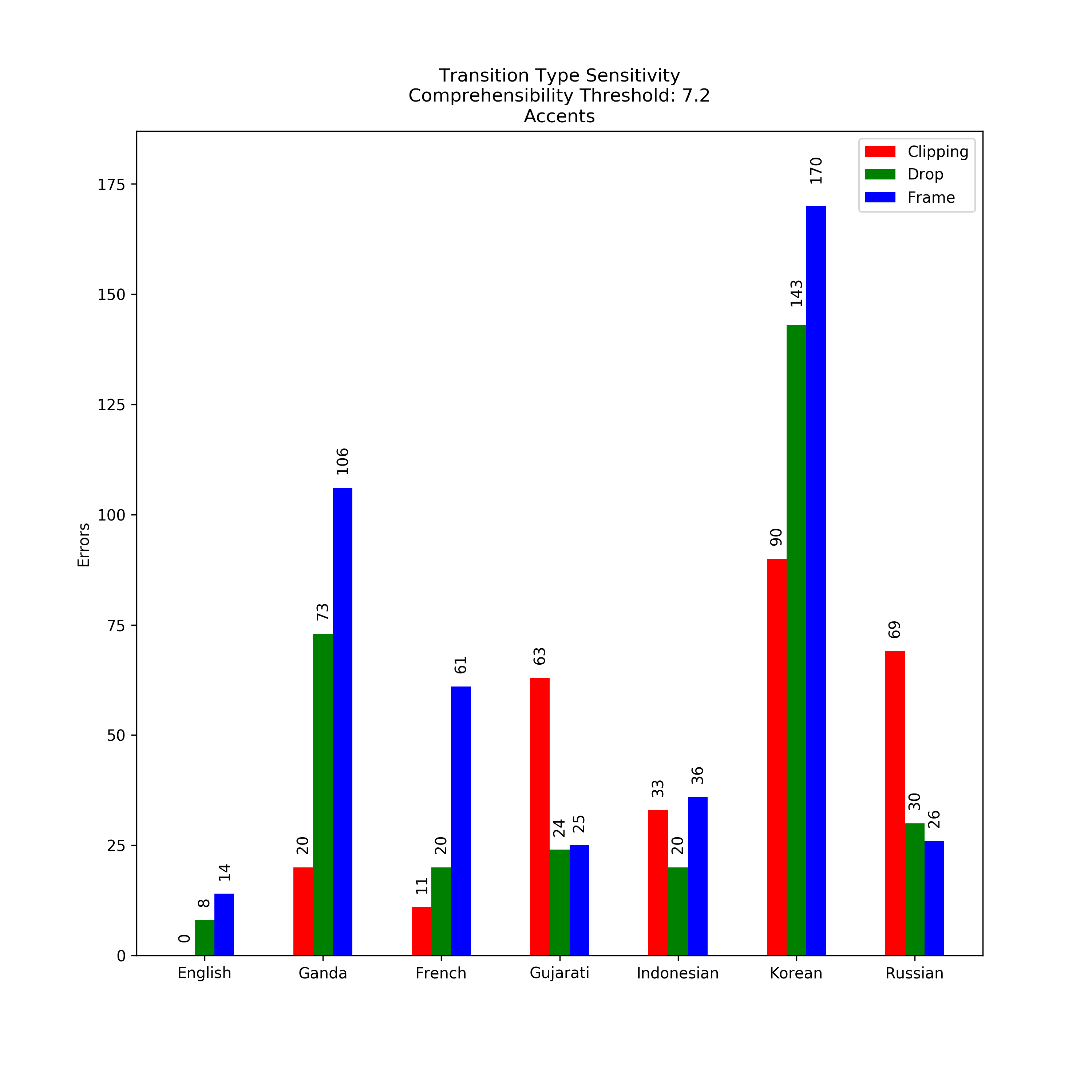} \\
{\bf (c)} & {\bf (d)}\\
\includegraphics[scale=0.25]{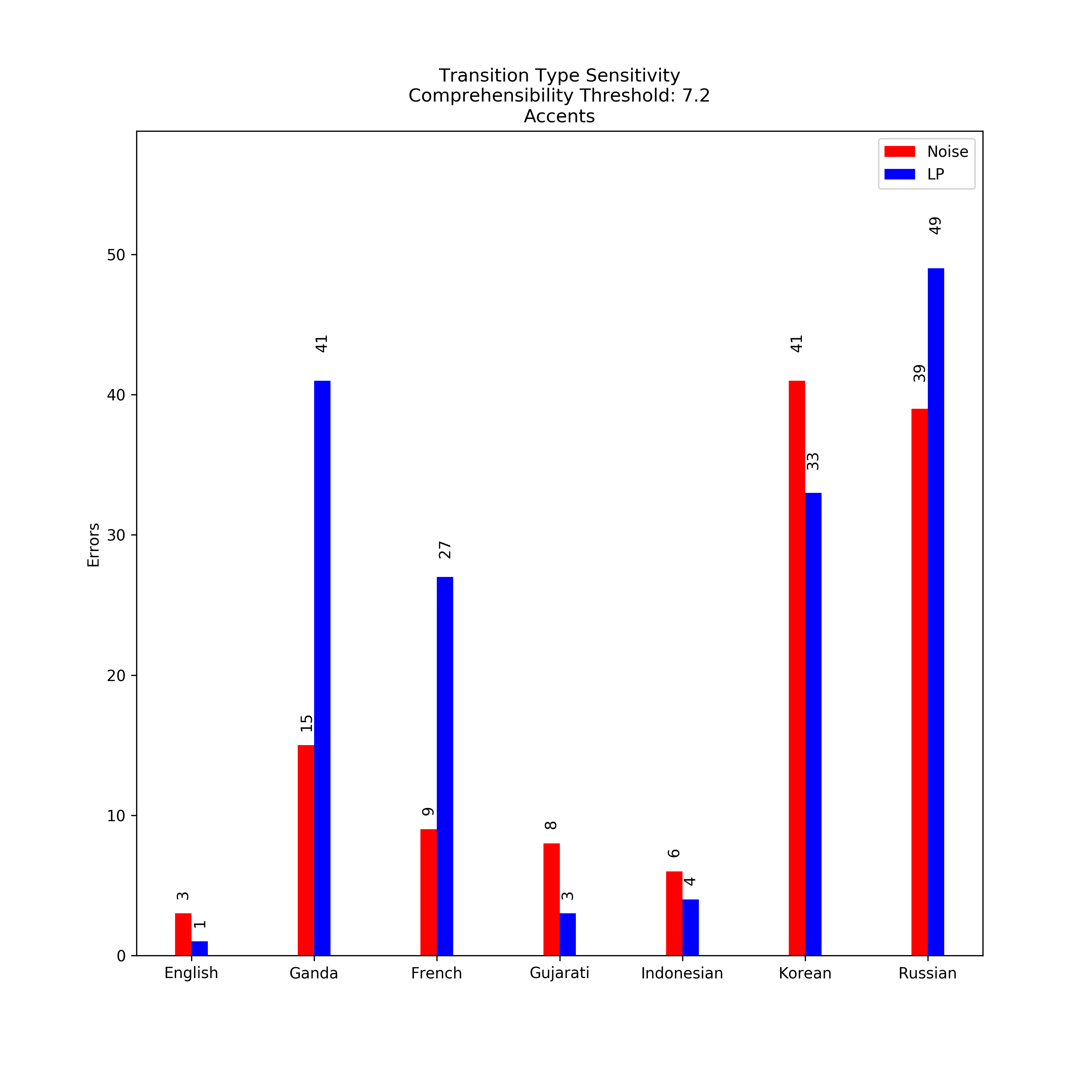} & 
\includegraphics[scale=0.25]{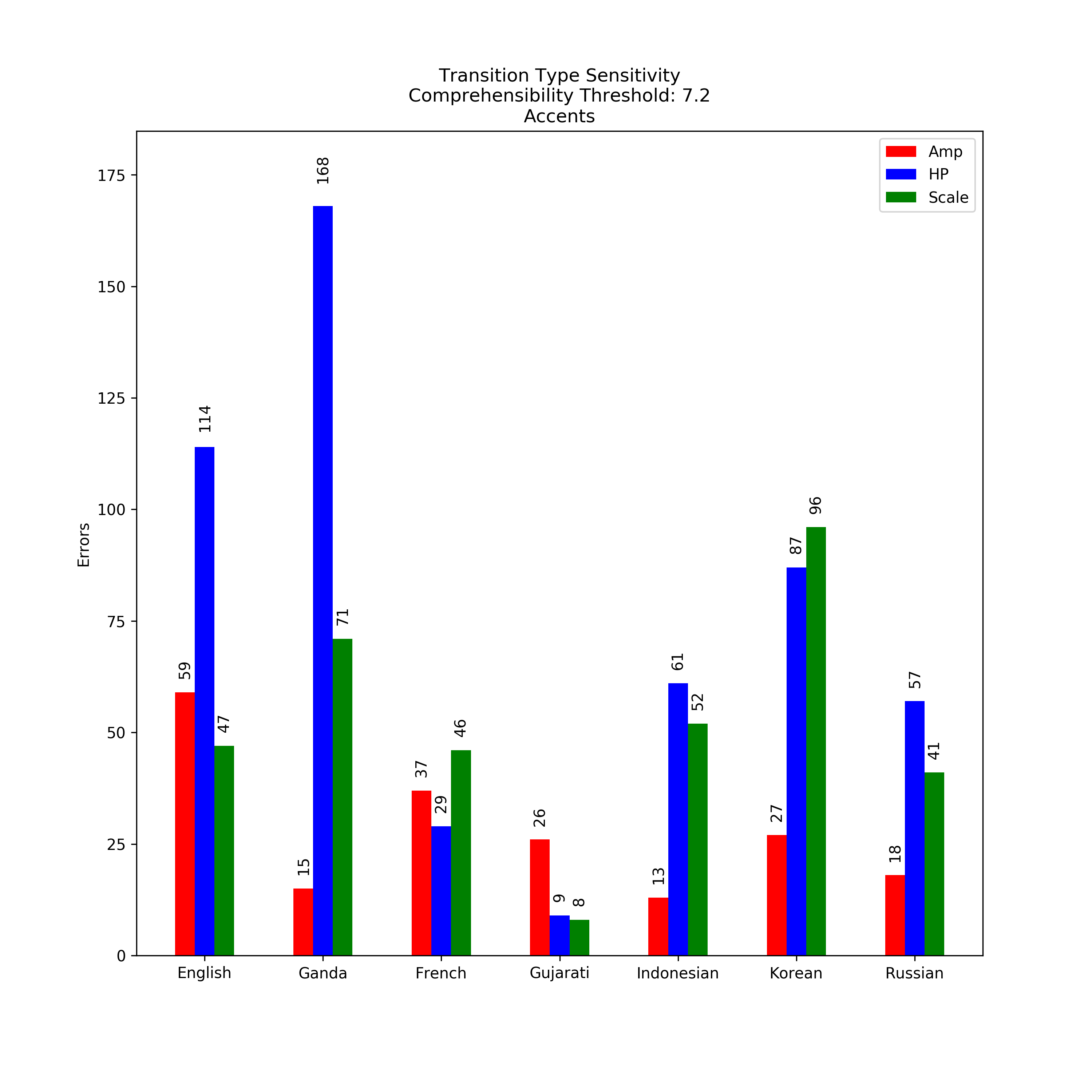} 
 \\
{\bf (e)} & {\bf (f)} \\
\end{tabular}}

\end{center}
\caption{Sensitivity analysis for the Accents dataset (with comprehensibility threshold 7.2 for transformations)}
\label{fig:accents-errors-comp-thresh-7-2}
\end{figure*}

\begin{figure*}[h]
\begin{center}
{
\begin{tabular}{cc}
\includegraphics[scale=0.25]{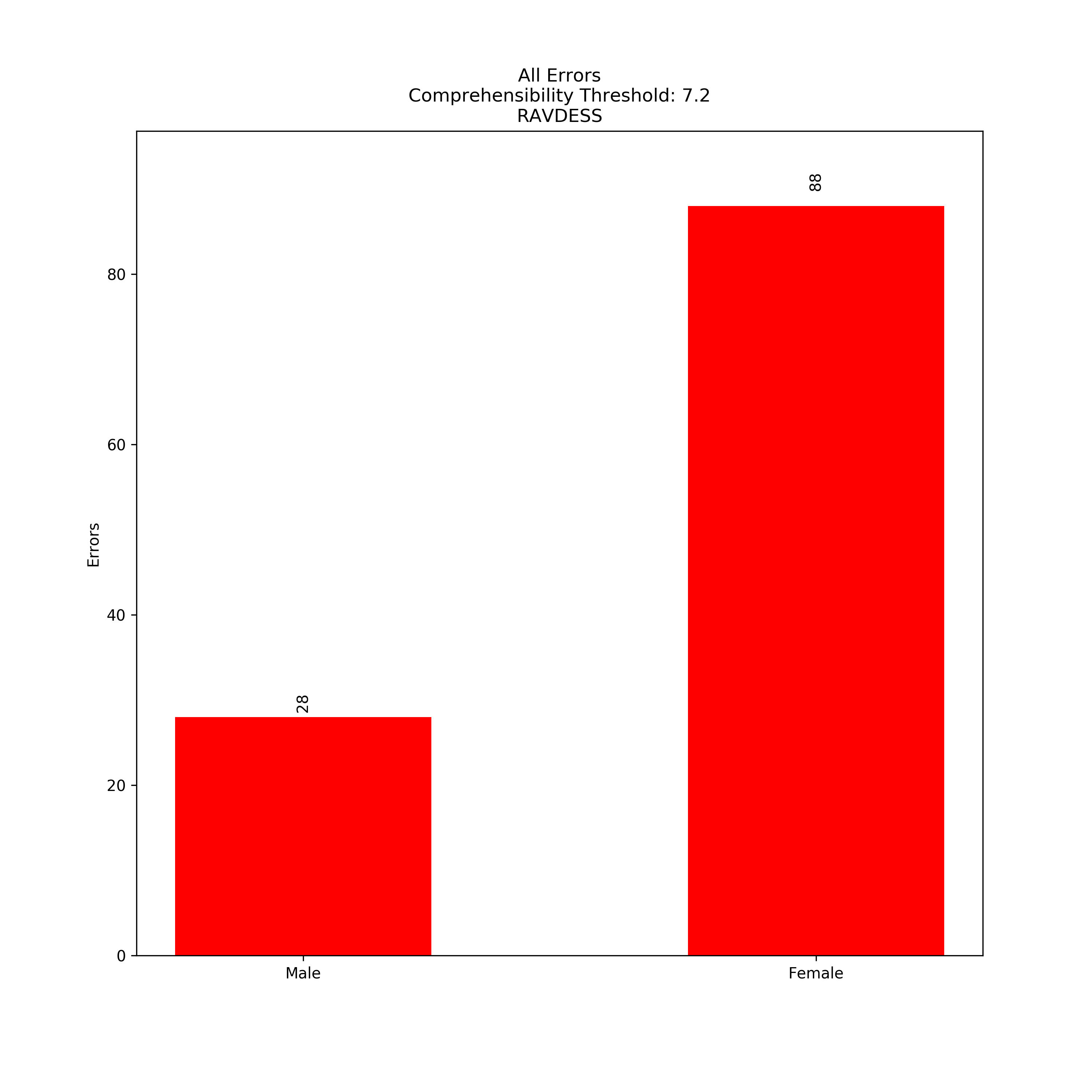} &
\includegraphics[scale=0.25]{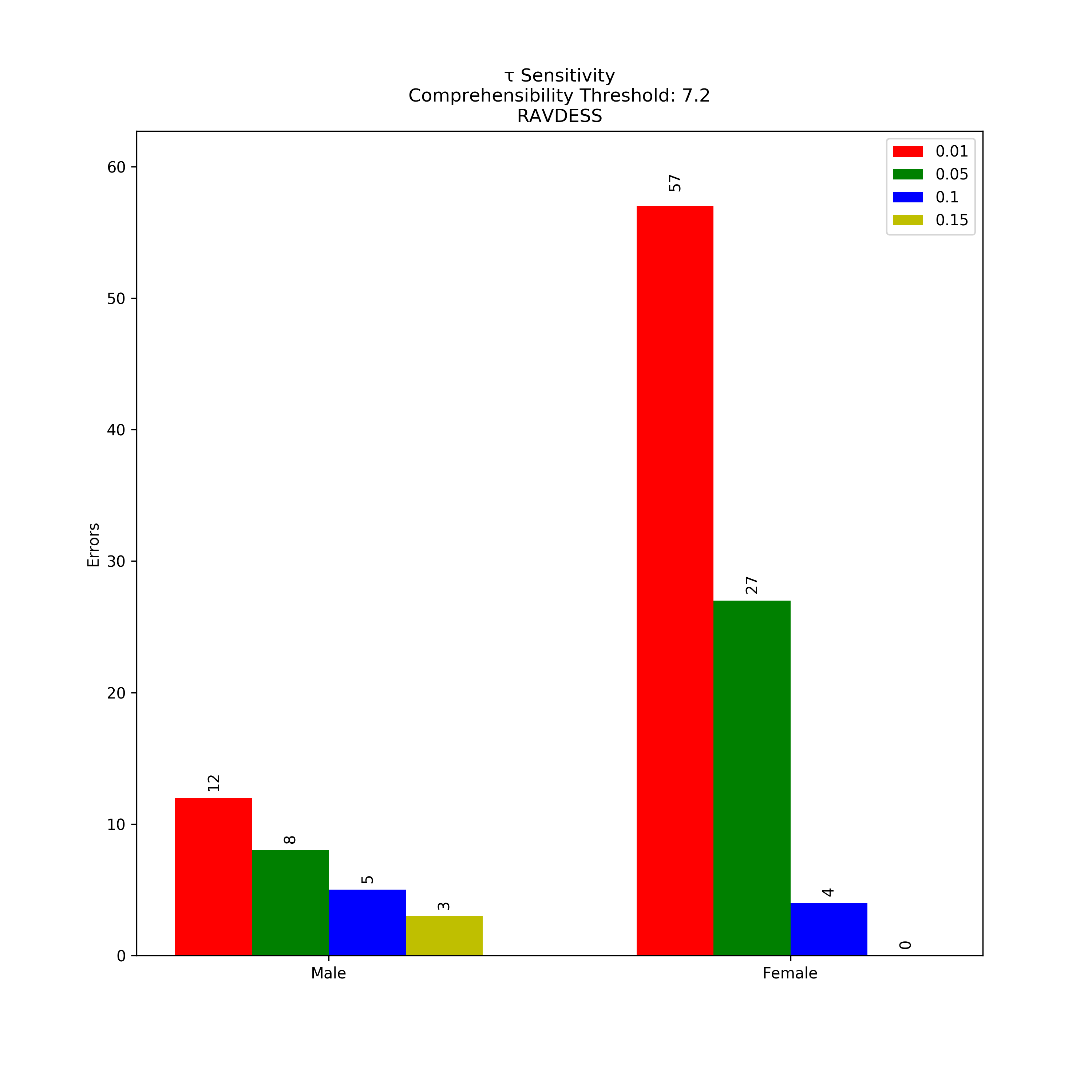} \\
{\bf (a)} & {\bf (b)}\\
\includegraphics[scale=0.25]{{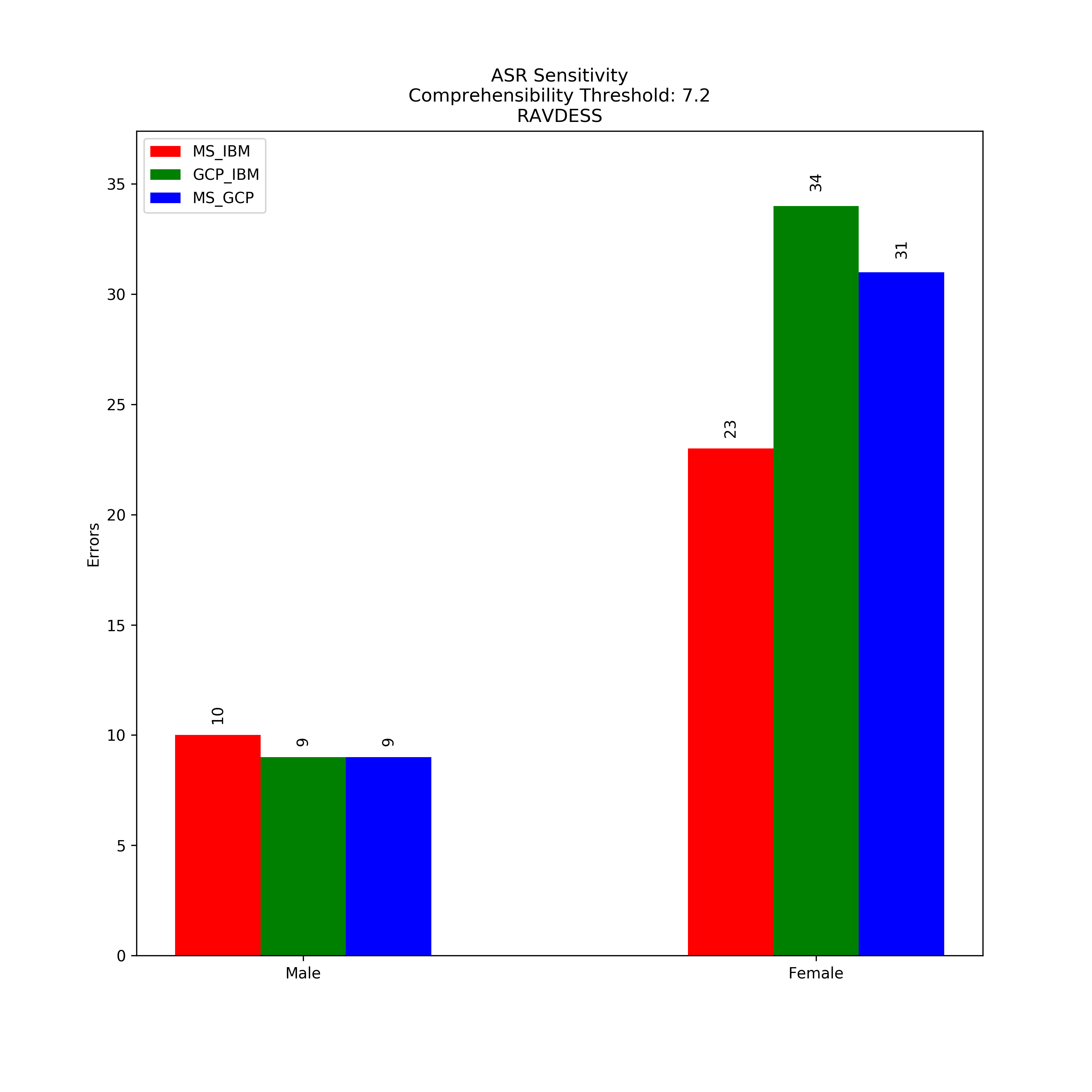}} &
\includegraphics[scale=0.25]{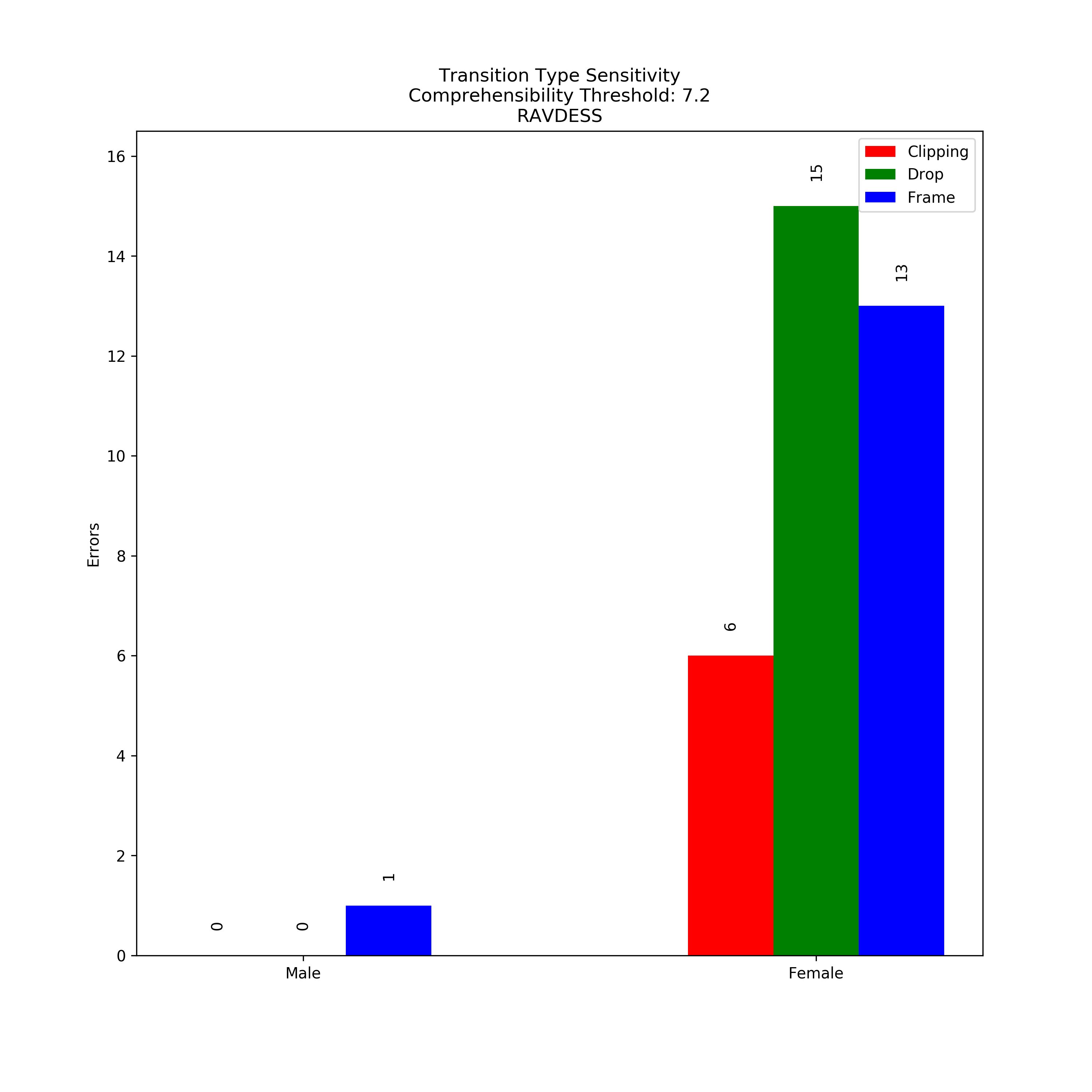} \\
{\bf (c)} & {\bf (d)}\\
\includegraphics[scale=0.25]{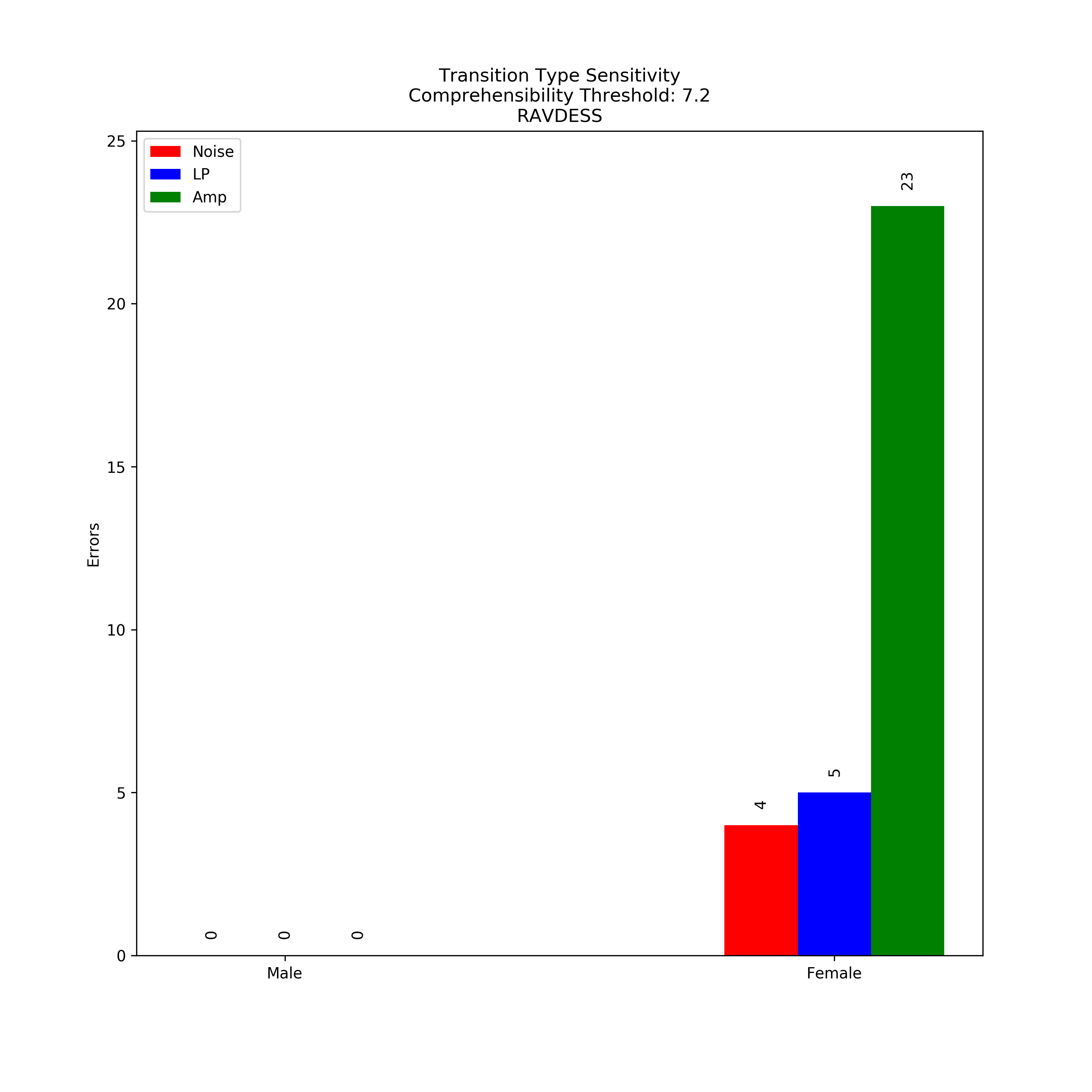} & 
\includegraphics[scale=0.25]{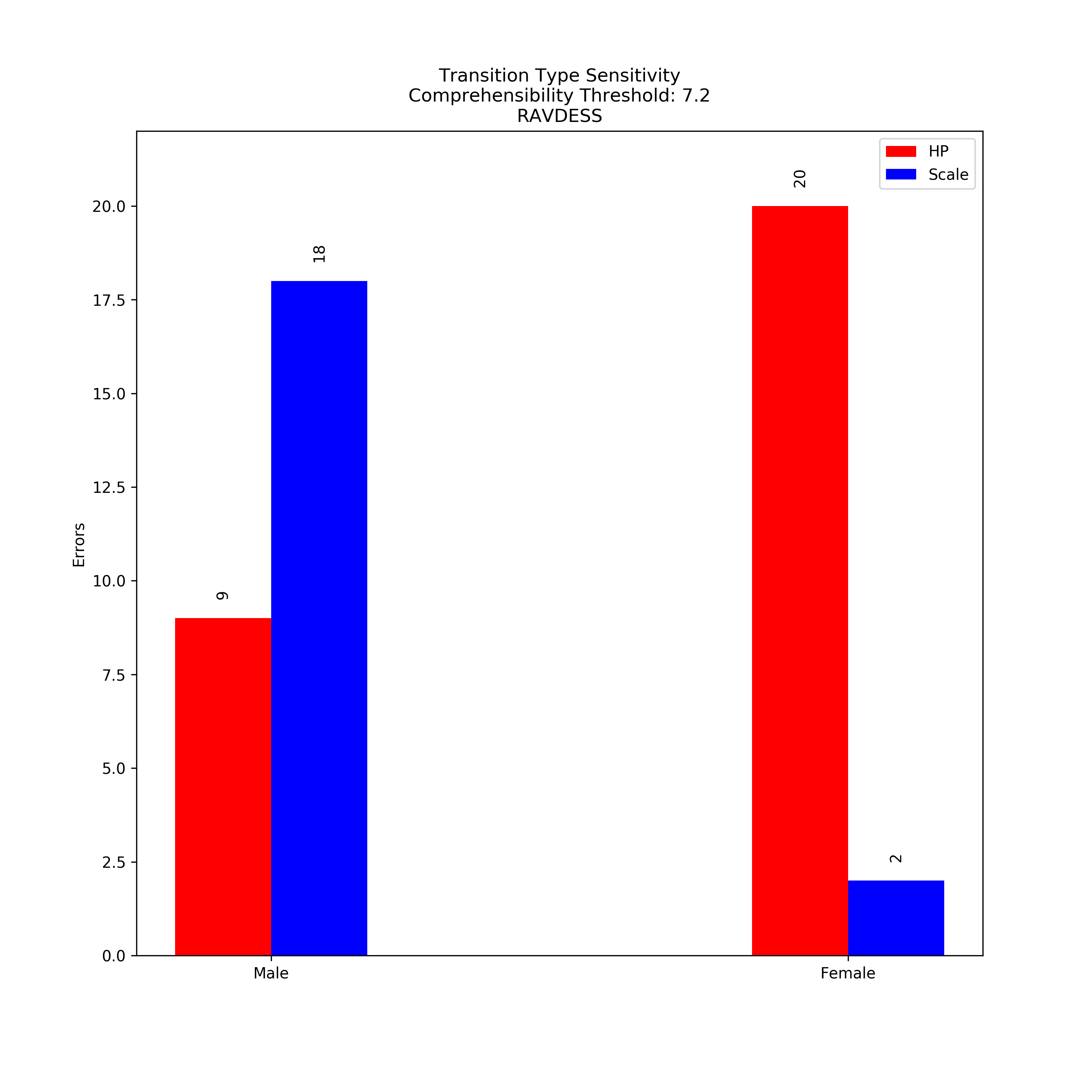} 
 \\
{\bf (e)} & {\bf (f)} \\
\end{tabular}}

\end{center}
\caption{Sensitivity analysis for the RAVDESS dataset (with comprehensibility threshold 7.2 for transformations)}
\label{fig:ravdess-errors-comp-thresh-7-2}
\end{figure*}

\begin{figure*}[h]
\begin{center}
{
\begin{tabular}{cc}
\includegraphics[scale=0.25]{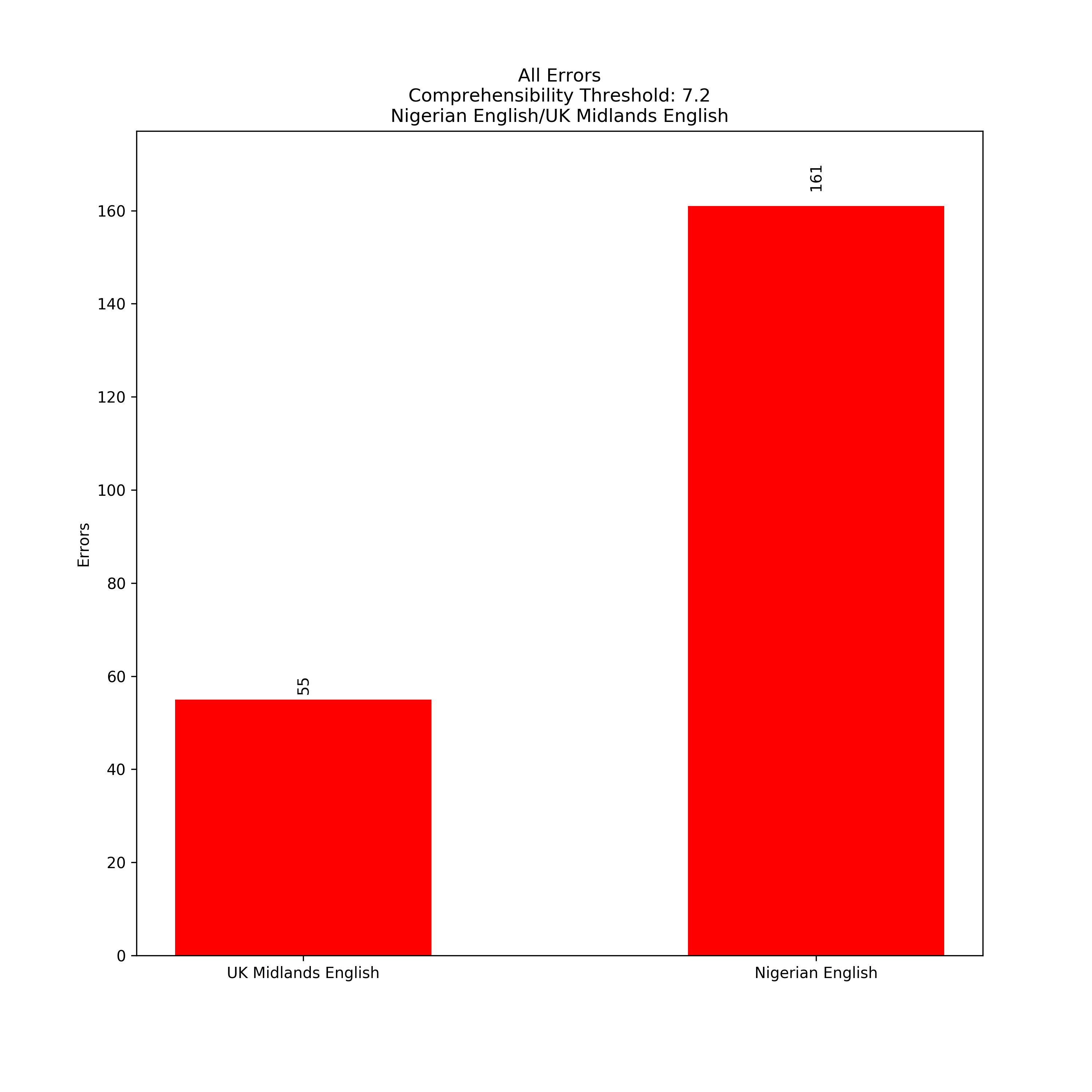} &
\includegraphics[scale=0.25]{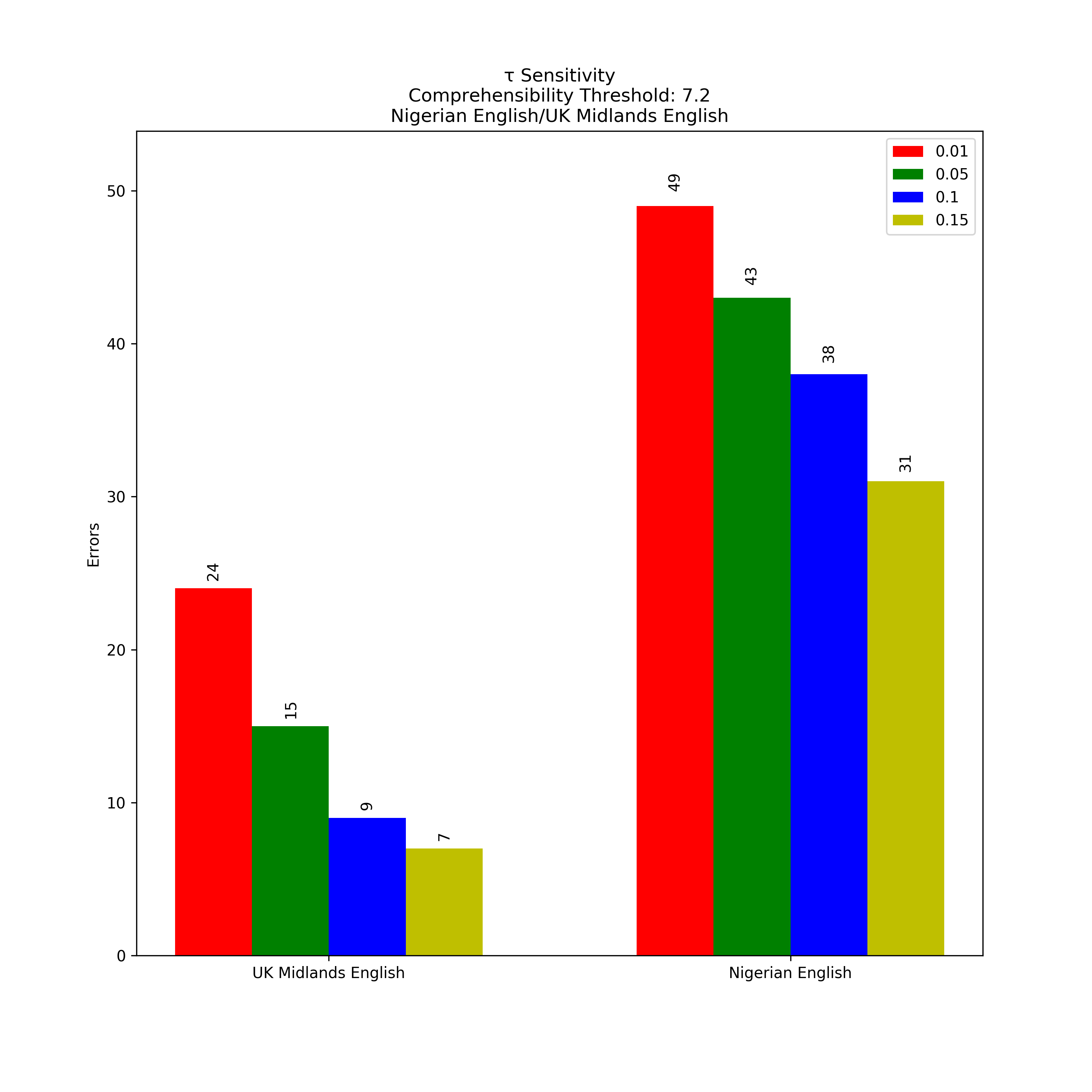} \\
{\bf (a)} & {\bf (b)}\\
\includegraphics[scale=0.25]{{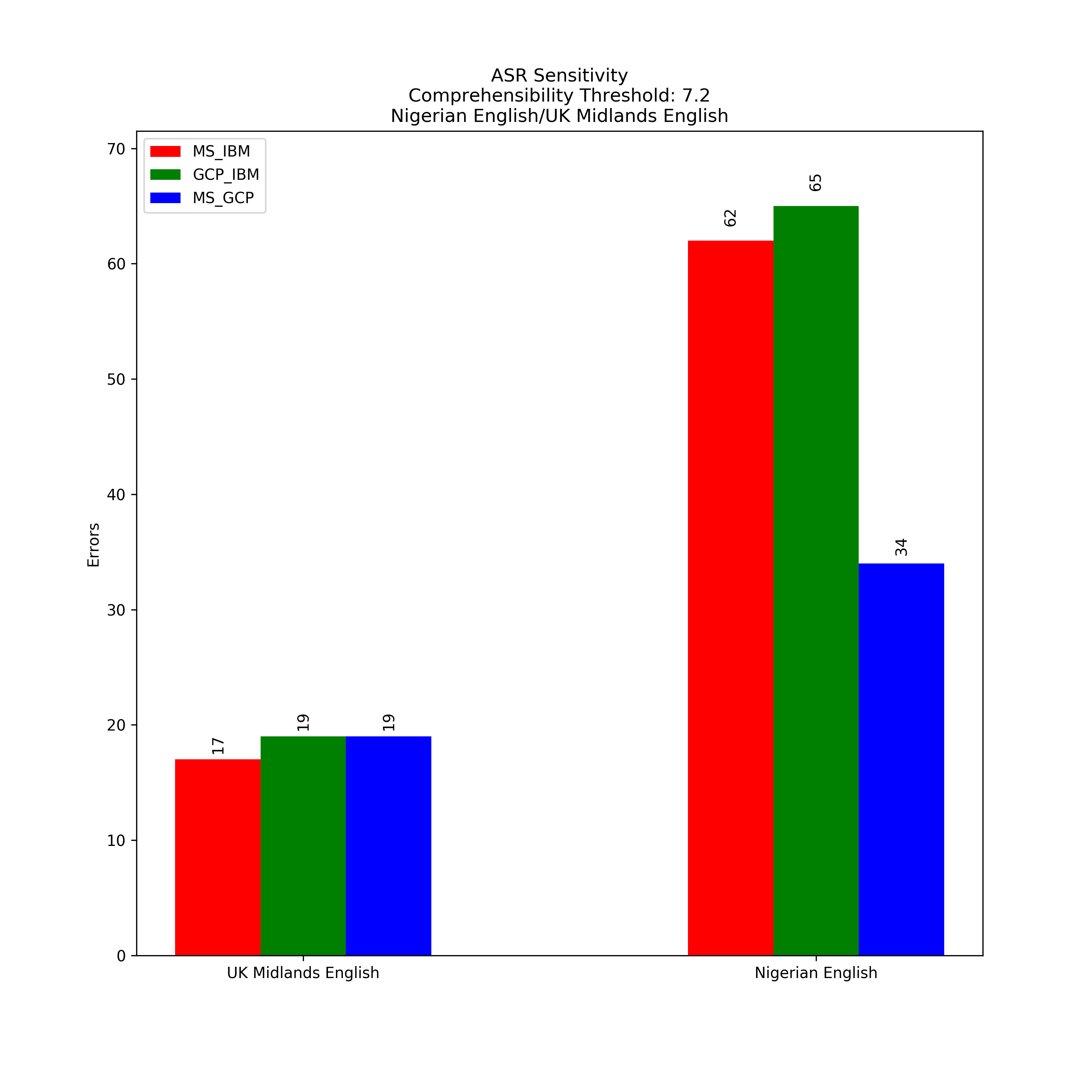}} &
\includegraphics[scale=0.25]{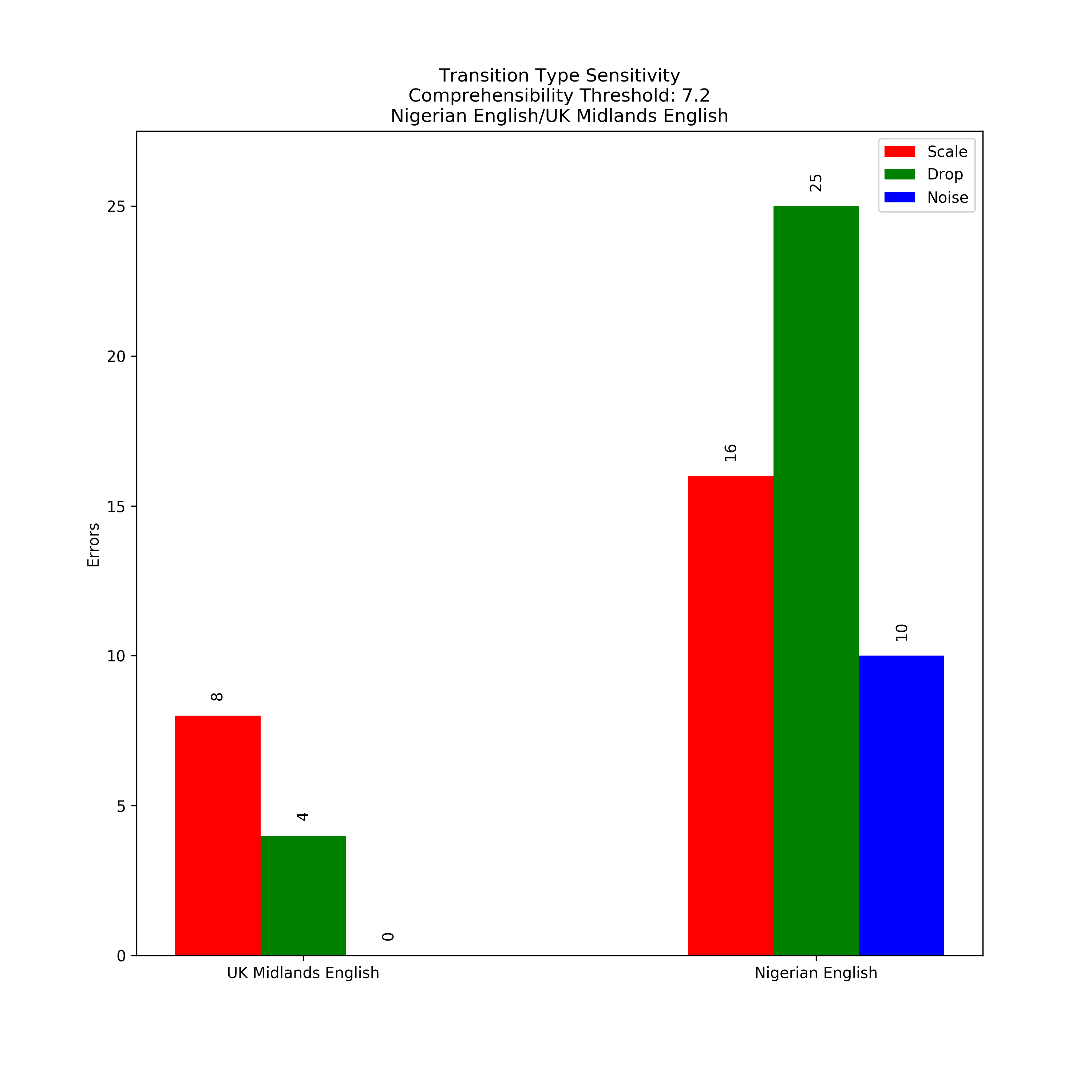} \\
{\bf (c)} & {\bf (d)}\\
\includegraphics[scale=0.25]{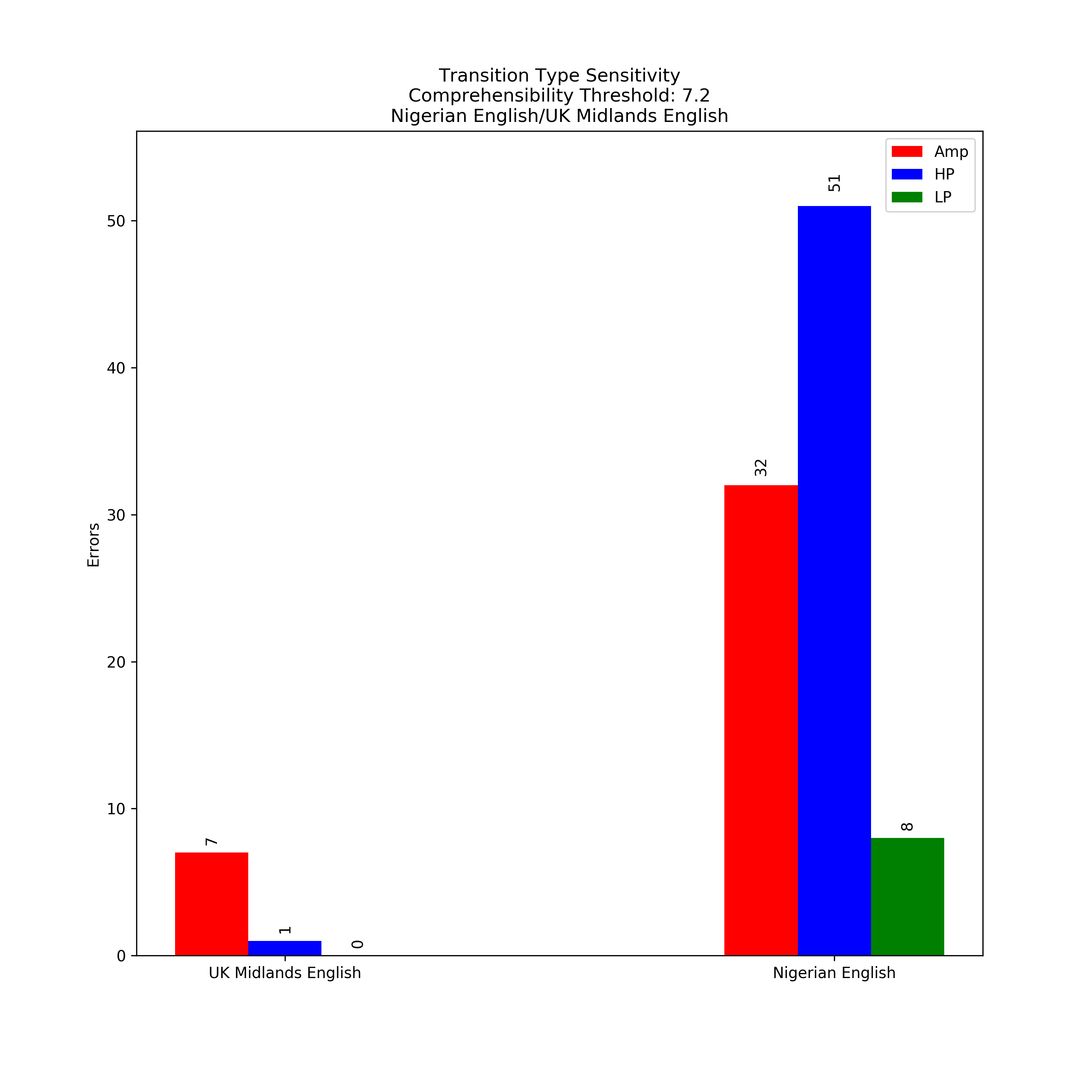} & 
\includegraphics[scale=0.25]{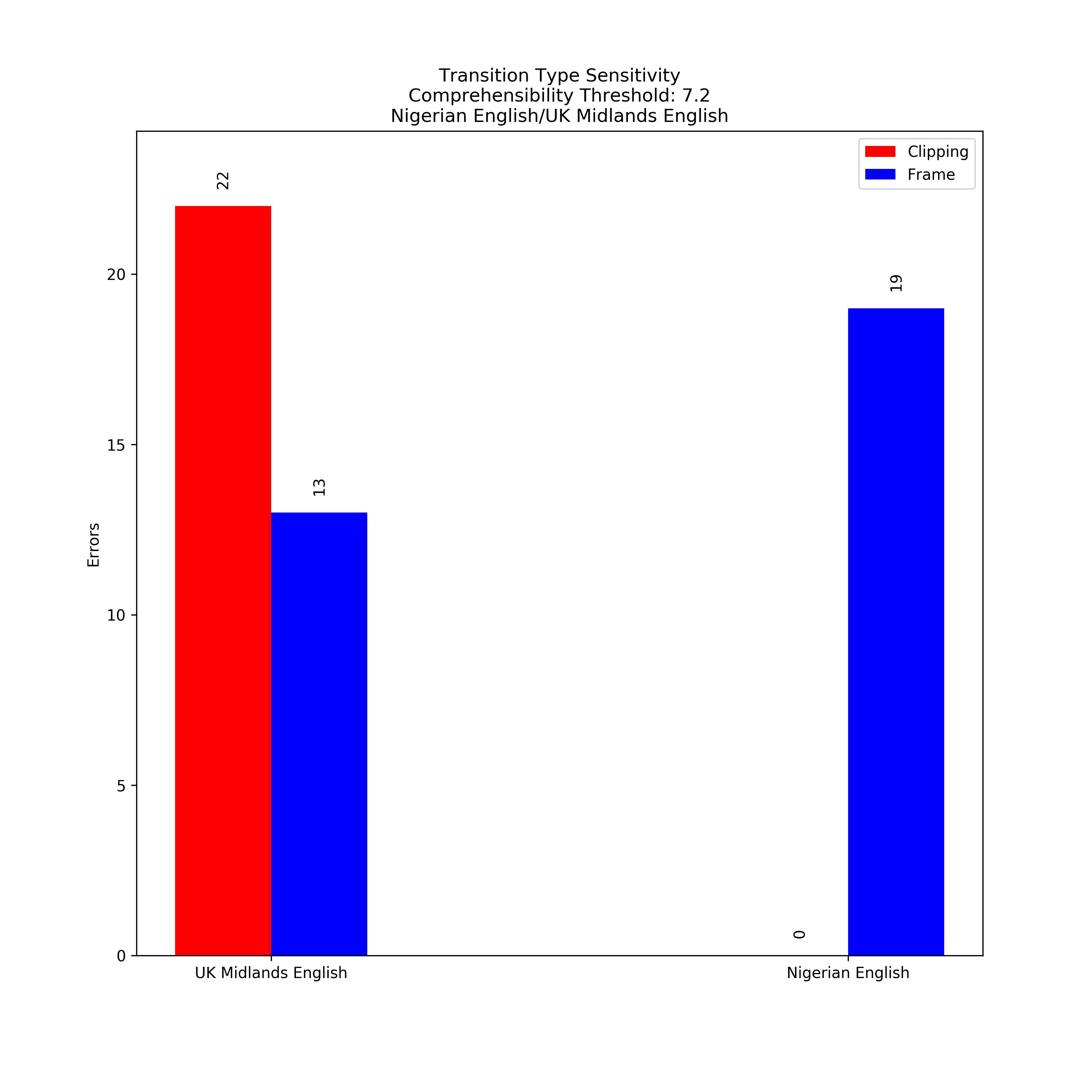} 
 \\
{\bf (e)} & {\bf (f)} \\
\end{tabular}}

\end{center}
\caption{Sensitivity analysis for the Nigerian English/UK Midlands English dataset (with comprehensibility threshold 7.2 for transformations)}
\label{fig:nigeria-midlands-errors-comp-thresh-7-2}
\end{figure*}

%

%
%
%
%
%
\end{document}